\documentclass[12pt]{iopart}
\usepackage{amstext}
\usepackage{xcolor}
\usepackage{mathptmx}
\usepackage{amsfonts}
\usepackage{graphicx}
\usepackage{subcaption}
\usepackage[super,sort&compress,comma]{natbib} 
\usepackage{hyperref, booktabs}
\usepackage{balance}



\graphicspath{ {FIGURES/} } 

\begin{document}

\title[Diverse Explanations in the Physical Sciences]{Diverse Explanations From Data-Driven and Domain-Driven Perspectives in the Physical Sciences}

\author{Sichao Li, Xin Wang \& Amanda Barnard}

\address{School of Computing, Australian National University, Canberra, Australia}
\ead{sichao.li@anu.edu.au}
\vspace{10pt}
\begin{indented}
\item[]October 2024
\end{indented}

\begin{abstract}
Machine learning methods have been remarkably successful in material science, providing novel scientific insights, guiding future laboratory experiments, and accelerating materials discovery. Despite the promising performance of these models, understanding the decisions they make is also essential to ensure the scientific value of their outcomes. However, there is a recent and ongoing debate about the diversity of explanations, which potentially leads to scientific inconsistency. This Perspective explores the sources and implications of these diverse explanations in ML applications for physical sciences. Through three case studies in materials science and molecular property prediction, we examine how different models, explanation methods, levels of feature attribution, and stakeholder needs can result in varying interpretations of ML outputs. 
Our analysis underscores the importance of considering multiple perspectives when interpreting ML models in scientific contexts and highlights the critical need for scientists to maintain control over the interpretation process, balancing data-driven insights with domain expertise to meet specific scientific needs. 
By fostering a comprehensive understanding of these inconsistencies, we aim to contribute to the responsible integration of eXplainable Artificial Intelligence (XAI) into physical sciences and improve the trustworthiness of ML applications in scientific discovery.
\end{abstract}

\section{Introduction}
\label{sec:intro}

The physical science studies rely heavily on the domain knowledge of scientists and often involve complex and computationally expensive simulations or economically expensive high-throughput experiments \cite{zhong2022explainable, kailkhura2019reliable, xu2020high}. The outcomes strongly depend on how comprehensively the search space is sampled, and this traditional domain-driven approach has clear limitations in prediction accuracy and efficiency. Machine learning (ML) approaches have received increasing attention as promising tools for materials research, particularly with the rise of deep neural networks (DNN) and numerous novel model structures proposed to enhance model predictability \cite{butler2018machine, schmidt2019recent, li2022inverse, liu2017materials}. 

ML models adopt a data-driven approach by analysing a large volume of historical data to derive new insights, based on various inputs such as physicochemical structure, state variables, or raw characterisation data, to reduce the dependency on domain knowledge or specific infrastructure. These models also demonstrate high accuracy in predicting a wide range of materials' physical, mechanical, optoelectronic, and thermal properties, such as crystal structure, melting temperature, formation enthalpy, and bandgap \cite{behler2016perspective, schmidt2019recent, schutt2014represent, yang2019establishing,Barnard2020SelectingML,Parker2021UnsupervisedSC}. Their success has driven the rapid adoption of ML models in scientific areas \cite{kailkhura2019reliable, wiens2018machine, carvalho2022artificial}.

\begin{figure}[h]
\begin{center}
\centerline{\includegraphics[width=.8\textwidth]{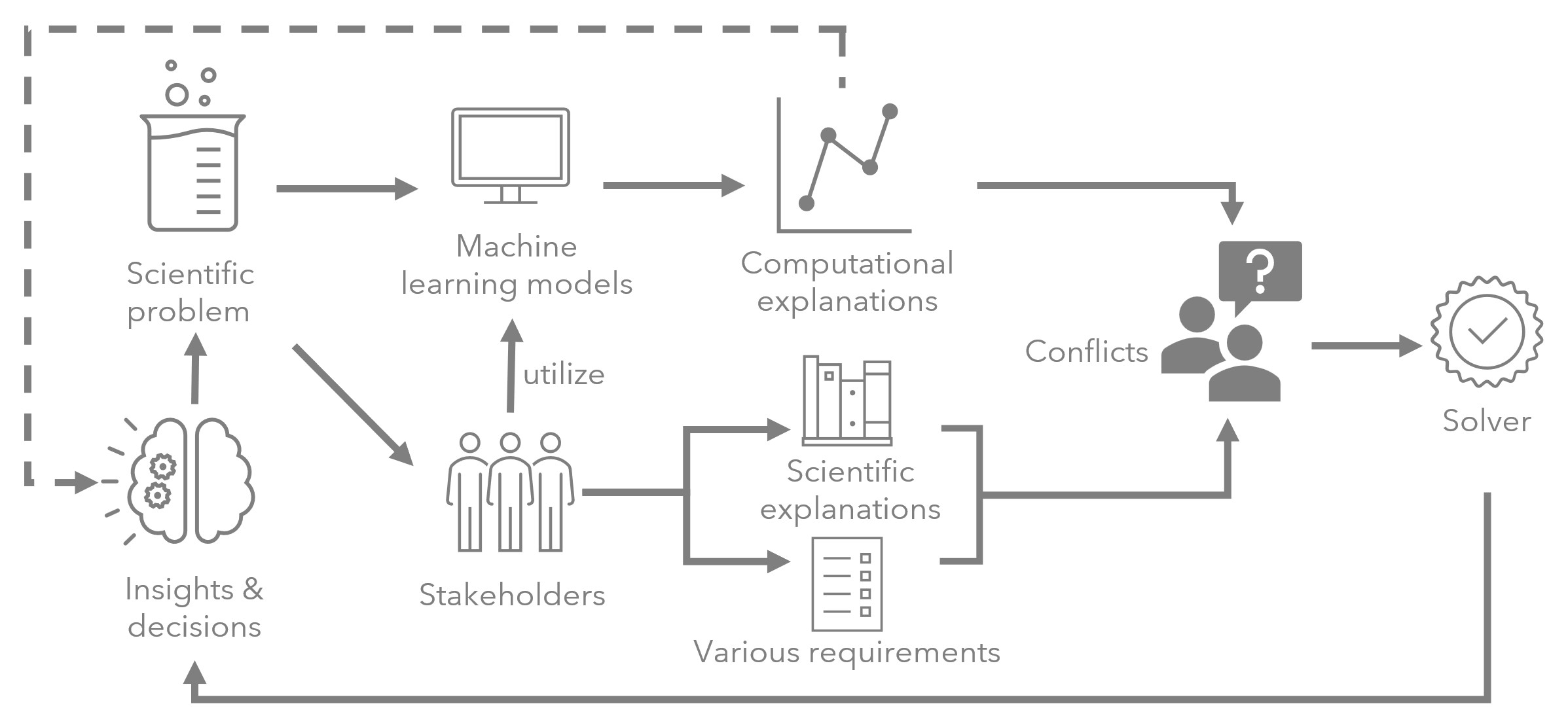}}
\caption{The potential conflicts from data-driven and domain-driven explanations in the decision-making process involving well-trained ML models. The dashed line denotes the conventional pipeline of XAI in the scientific domain, where stakeholders utilise ML models and retrieve data-driven explanations to analyse results. In practice, it is common that data-driven explanations conflict with domain knowledge, misleading researchers, and requiring a new approach. }
\label{fig:pipeline}
\end{center}
\end{figure}

However, the most accurate ML models, such as DNNs, are frequently difficult to explain and are often referred to as ``black-boxes''. The lack of explainability of ML methods has hindered their potential impact in many domains \cite{varshney2017safety}. Methods of explainability/interpretability for ML models are intensively studied in the field of computing, and there is an increasing demand for applying XAI to ML predictions in science \cite{jimenez2020drug, huang2023explainable, zhong2022explainable, barnard2023importance, barnard2022explainable}. For instance, a typical scenario where a new material property is predicted by an ML model based on the structure can be challenging to optimise and translate to manufacturers without understanding how specific material characteristics influence the prediction. For this reason, many scientists find black-box ML models difficult to trust.


The problem is exacerbated by the fact that recent studies have demonstrated that many different ML models can perform similarly on the same dataset, even with different functional forms~\cite{rudin2019stop, sichao2023vtf, fisher2019all}. As a result, ML models with comparable effectiveness can offer diverse explanations for the same task and an intuitive question easily arises: which model should we trust?
Many researchers seek to identify which samples or features are important in describing the model effects \cite{li2023exploring, hsu2022rashomon}. However, when faced with the existence of multiple ``equally-good'' models, training and explaining with a single model becomes problematic, and results in inconsistencies between interpretability and established scientific principles \cite{reichstein2019deep, roscher2020explainable}.  Figure \ref{fig:pipeline} illustrates the potential conflicts between data-driven and domain-driven explanations in the decision-making process involving well-trained ML models, where the term ``well-trained'' models refer to ML models that demonstrate promising performance on their respective tasks. The performance of these models is crucial, as the reliability of explanations derived from a model is contingent on its predictive capability.

\begin{figure}[t!]
    \centering
    \begin{subfigure}[t]{0.24\textwidth}
         \centering
         \includegraphics[width=\textwidth]{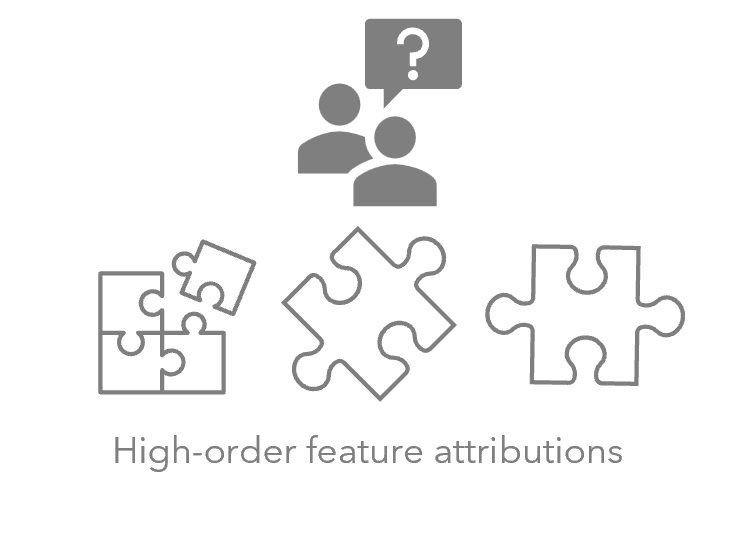}
         \label{fig:three sin x}
     \end{subfigure}
     \hfill
          \begin{subfigure}[t]{0.25\textwidth}
         \centering
         \includegraphics[width=\textwidth]{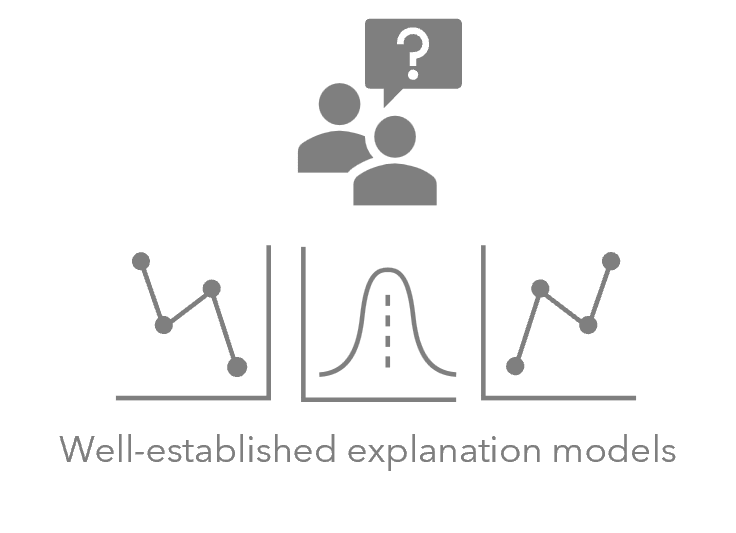}
         \label{fig:three sin x}
     \end{subfigure}
     \hfill
     \begin{subfigure}[t]{0.24\textwidth}
         \centering
         \includegraphics[width=\textwidth]{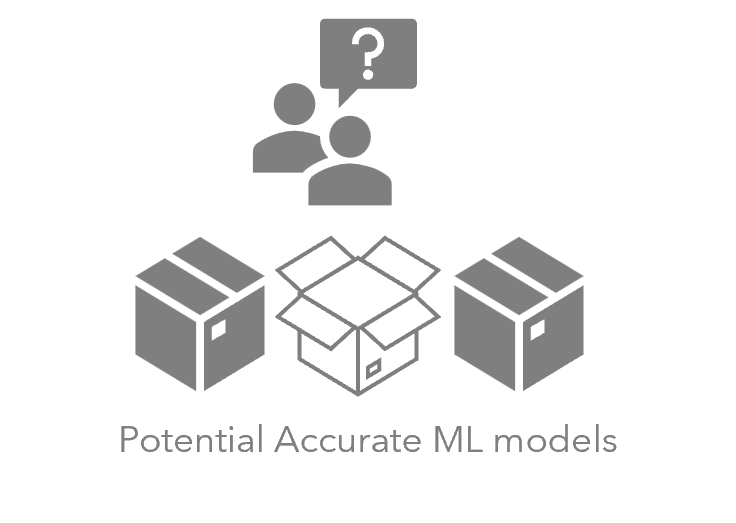}
         \label{fig:three sin x}
     \end{subfigure}
     \hfill
    \begin{subfigure}[t]{0.24\textwidth}
         \centering
         \includegraphics[width=\textwidth]{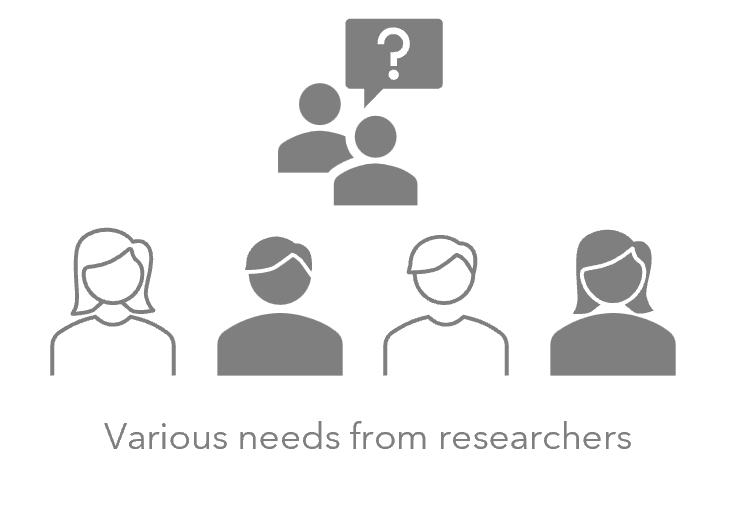}
         \label{fig:three sin x}
     \end{subfigure}
     \vspace{-.5cm}
    \caption{Visualisation of four sources of inconsistent explanations.}
\label{fig:challenges-all}
\end{figure}

In this Perspective, we argue that, even though there are a large number of XAI studies in scientific domains, trust has yet to be established.  We focus on well-trained models to ensure that the explanations we analyse are based on models with predictions that can be reasonably trusted.
We identify and explore diversities between data-driven and domain-driven explanations in well-trained ML models from different perspectives, shown in Fig. \ref{fig:challenges-all}. Through three case studies in materials science, nanotechnology and molecular property prediction, we highlight how different models, explanation methods, levels of feature attribution, and stakeholder needs can result in varying interpretations for identical tasks. This illuminates the multifaceted nature of explanations in scientific applications, and emphasises the crucial role of scientists in guiding the interpretation process.

\section{Background and Concepts}

\subsection{Explainability and Interpretability in Science}
Explainability and interpretability are closely related concepts in XAI, often used interchangeably to describe the ability of humans to understand model predictions. In scientific contexts, these concepts are particularly crucial as they bridge the gap between data-driven ML models and domain-driven scientific understanding \cite{linardatos2020explainable, lipton2018mythos,kim2016examples, miller2019explanation, sichao2023vtf, imrie2023multiple, adadi2018peeking, doshi2017towards,Liu2023TheER}. 
In scientific applications, the need for explainable and interpretable ML models is crucial \cite{freitas2014comprehensible, rudin2019stop}. Scientists not only need accurate predictions but also require insights into the underlying mechanisms. This aligns with the scientific method, where understanding the reason behind a prediction is as important as observing the phenomenon itself.

\subsection{The Foundation of Trust: Performance and Explanation}
Trust in ML models in scientific and technology is built on two pillars: performance and explanation \cite{rudin2019stop}. 
Performance metrics such as Mean Absolute Error (MAE), coefficient of determination (${R^2}$), and Area Under the Receiver Operating Characteristic curve (ROC-AUC) quantify how well a model predictions aligns with ground truth. However, these metrics alone do not provide insights into the scientific validity of the underlying decision-making process.
The main challenge lies in the fact that high-performing models, especially complex neural networks, are often difficult to interpret, creating a trade-off between performance and explainability. 


\subsection{Explanation Categories in Science: Domain-Driven and Data-Driven}
Explanation methods in ML can be categorised in many ways, e.g., feature-based vs. instance-based, post-hoc vs. ante-hoc, or output-focused vs. model-focused \cite{gilpin2018explaining, guidotti2018survey, adadi2018peeking, linardatos2020explainable, zhong2022explainable, roscher2020explainable, imrie2023multiple, roscher2020explainable}. In this Perspective, we focus on two primary categories of explanations in scientific ML: domain-driven and data-driven approaches. This categorisation provides a framework for understanding the sources of diverse explanations in scientific applications of ML.

\paragraph*{Data-Driven Explanations}primarily rely on patterns and relationships discovered in the training data. These approaches are commonly used when integrating ML methods into scientific domains, as they can uncover insights directly from large volumes of data. Data-driven explanations often employ post-hoc interpretation methods on complex, well-performing ML models \cite{nauta2023anecdotal, zhong2022explainable, gola2018advanced, pankajakshan2017machine, 9007737}. Data-driven explanations offer significant advantages, such as the ability to uncover hidden patterns in big data. The flexibility of data-driven methods also support a variety of application without the need for building complex, domain-specific ML models. However, their limitation lies in the potential to produce explanations that, while mathematically sound, may lack clear scientific interpretation or may even conflict with established theories.

\paragraph*{Domain-Driven Explanations}leverage expert knowledge and established scientific principles to construct domain-specific ML models or interpret ML model outputs. These explanations aim to align model behavior with existing theories and physical laws. For instance, attention-based models such as CrabNet \cite{wang2021compositionally} can provide element-wise contributions to property predictions, aligning with domain knowledge about elemental influences.  The strength of domain-driven explanations is their scientific consistency and interpretability within the context of existing knowledge. They often result in explanations that are more readily accepted and understood by domain experts. However, they can be limited by their dependence on both current scientific understanding and ML modeling, which may inadvertently constrain the model performance, and ability to discover novel patterns or relationships not yet recognised in the field \cite{rudin2019stop}.

\paragraph*{Interplay Between Approaches}does not mean that domain-driven and data-driven explanations are mutually exclusive categories. In practice, many effective explanation methods incorporate elements of both approaches. Inherently interpretable models can be data-driven while incorporating domain knowledge, such as physics-informed neural networks that embed known physical laws into their architecture. Conversely, complex black-box models can be designed with domain-specific constraints or analysed using explanation methods that align with domain understanding.  However, balancing these approaches can lead to explanations that are both scientifically consistent and capable of revealing novel insights, potentially driving forward our understanding in sciences.



\subsection{Explanation Inconsistency}
Scientific consistency \cite{karpatne2017theory} was introduced as an essential component for model learning in 2017, among the fundamental requirements for producing trustworthy outcomes in scientific applications \cite{reichstein2019deep}. This means that the results obtained from ML models must be consistent with established scientific principles \cite{roscher2020explainable, Zhang2011AvoidingSC}. This concept bridges the gap between data-driven predictions and domain knowledge. Instead of scientific consistency, which is subjective and hard to maintain, we explore scientific inconsistency through the lens of data-driven and domain-driven explanations. Our focus is on feature-based explanations in human-understandable terms, providing visualisations of feature importance rankings to illustrate diverse explanations.

\subsection{Feature-Based Explanations in ML}
Feature-based explanations play a crucial role in interpreting model behavior in human-understandable terms \cite{nauta2023anecdotal,zhong2022explainable, gola2018advanced, pankajakshan2017machine, 9007737}. In this Perspective, we focus specifically on feature importance as a key aspect of these explanations.

\paragraph{Feature Importance Measurements}are of the most commonly used methods for explaining ML models is feature importance ranking, derived from input-output relationships~\cite{pankajakshan2017machine, 9007737}. In regression tasks, feature importance is calculated based on how much a feature contributes to the predicted value; while in classification, feature importance is calculated based on the contribution to each class. Although complex neural networks are hard to interpret, some simpler models can inherently provide explanations. For example, a decision tree is a rule-based classification algorithm that splits at each node according to metrics such as the Gini index, which is computed as:
\begin{equation*}
\text{Gini} = 1 - \sum_{i=1}^{C} p_i^2
\end{equation*}
where $p_i$ is the current percentage of class $i$ and $C$ is the number of classes. The difference of Gini index at each split is calculated as the difference between the parent node and the weighted average of child nodes, such that:
\begin{equation*}
    \Delta \text{Gini} = \text{Gini}_{\text{parent}} - \left(\frac{N_{\text{left}}}{N} \times \text{Gini}_{\text{left}} + \frac{N_{\text{right}}}{N} \times \text{Gini}_{\text{right}}\right)
\end{equation*}
where $N$ denotes the number of instances in the node. Decision trees\cite{breiman2017classification} iterate through all features and possible values to find the one that maximises the Gini index difference and uses it for the split. After training, feature importance can be calculated as the sum of Gini index differences among all nodes where the feature is used for split, and divided by the sum of Gini index differences among all features. 

More sophisticated methods such as random forests and XGBoost\cite{chen2015xgboost} combine decision trees with additional mechanisms such as regularisation. The importance of features is calculated by the average gain across all nodes of all trees where the feature is used for splits. In the absence of these intrinsic model explanations, some well-established methods can offer universal explanations for both simple and complex models. The explanation methods highlighted in this study are summarised in Table \ref{tab:explanation_methods}.

\begin{table*}[h]
\centering
\footnotesize
\begin{tabular}{p{0.25\textwidth}p{0.45\textwidth}p{0.13\textwidth}}
\toprule
\textbf{Explanation Methods} & \textbf{Description} & \textbf{Usage} \\ \midrule
Shapley Additive Explanations (SHAP)\cite{lundberg2017unified} & A game theory-based method that computes the marginal contribution of each feature by calculating the Shapley value. & Case study 1 Sec.\ref{subsec:nano} \\
Permutation Importance (PI)\cite{breiman2001random} & Evaluate feature importance by invalidating features and measuring the difference in model performance. & Case study 1 Sec.\ref{subsec:nano}\\
Local Interpretable Model-agnostic Explanations (LIME)\cite{ribeiro2016should} & Generates new data instances near a specific instance and trains a simple model on these to obtain local explanations. & Case study 3 Sec.\ref{subsec:bace}\\
Integrated Gradients (IG)\cite{sundararajan2017axiomatic} & Computes feature importance by integrating the gradients of the model's output with respect to the input features. & Case study 1 in Sec.\ref{subsec:nano}\\
Feature Interaction Score (FIS)\cite{li2023exploring} & Measures the performance change of feature attributions from the baseline, quantifying different levels of feature attributions. & Case study 2 Sec.\ref{sec:aflow}\\
Connection Weights (NN) \cite{beck2018neuralnettools, olden2004accurate} & Analyses the weights of connections in neural networks to determine feature importance. & Case study 1 in Sec.\ref{subsec:nano}\\
Gini Importance (Decision Trees)\cite{breiman2017classification} & Calculates feature importance in decision trees based on the Gini impurity criterion. & Case study 3 Sec.\ref{subsec:bace}\\
Average Gain (XGBoost) \cite{chen2015xgboost} & Computes feature importance in XGBoost models based on the average gain across all splits where the feature is used. & Case study 3 Sec.\ref{subsec:bace}\\
\hline
\end{tabular}
\caption{Summary of explanation methods highlighted in this perspective.}
\label{tab:explanation_methods}
\end{table*}


\subsubsection{Feature Attribution Scores}can be used to ensure consistency when comparing feature importance across different models and methods \cite{ribeiro2016model, ribeiro2016should, sundararajan2017axiomatic, janizek2021explaining}. This approach allows for a fair comparison of feature importance in a broader context, especially when dealing with complex feature interactions. Feature attribution refers to the process of assigning importance or contribution values to input features with respect to a model prediction, indicating the importance of a feature. In particular, feature interaction score (FIS) is a convenient way quantify different levels of feature attributions \citep{fisher2019all, li2023exploring, li2024practical, dong2020exploring}, such that:
\begin{equation*}
    \varphi_{i}(f_{ref}) = \mathbb{E}[L(f_{ref}(\mathbf{X}_{\setminus s}), \mathbf{y})] - \mathbb{E}[L(f_{ref}(\mathbf{X}), \mathbf{y})]
\end{equation*}
where $\mathbf{X}_{\setminus s}$ denotes the input matrix when the feature of interest is replaced by an independent variable. This method measures the performance change of feature attributions from a baseline $L_{ref} = \mathbb{E}[L(f_{ref}(\mathbf{X}), \mathbf{y})]$. 

The ``level'' of feature attribution indicates the complexity of the relationships being explained.  Individual feature importance is referred to as first-order feature importance, while the importance of relationships between two features is referred to as second-order feature interaction. Higher-order interactions involve combinations of three or more features, resulting in higher-order feature interactions. Similarly, higher-order feature attributions can be represented by:
\begin{equation*}
    \varphi_{I}(f) =  \mathbb{E}[L(f_{ref}(\mathbf{X}_{\setminus I}), \mathbf{y})] - \mathbb{E}[L(f_{ref}({X}), \mathbf{y})]
\end{equation*}
where $I$ is a set of features, formalised as $|I| > 1$.
In practice, one can permute the features of interest multiple times to achieve a similar measurement \citep{datta2016algorithmic}.  With this in mind, the feature attribution score is defined as the difference between the loss change of replacing features simultaneously and the sum of the loss change of replacing multiple features individually:  
\begin{equation*}
FIS_{I}(f_{ref}) = \varphi_{I}(f_{ref})-\sum_{i \in I}\varphi_{i}(f_{ref}).
\label{eq:fis_definition}
\end{equation*}
For common feature importance, $FIS_{s}(f_{ref}) = \varphi_{s}(f_{ref})$ and the loss can be approximated by the model performance metrics mentioned above.  

Both first-order and second-order explanations have value in the physical sciences, where importances and interactions can be underpinned or anticipated by domain knowledge.  These sorts of explanations are often intuitive for researchers, but must be explicitly calculated for ML models, leading to potential inconsistencies.

\subsubsection{Implications for Scientific ML}
The ability to calculate and compare feature attribution scores across different levels, models, and methods allows researchers to identify and analyse various types of inconsistencies in explanations:
\begin{itemize}
    \item Data-driven Inconsistencies: By quantifying feature importance using different data-driven approaches (e.g., SHAP, PI, and LIME), researchers can directly compare and contrast explanations derived from the same data but different methods.
    \item Domain-driven Inconsistencies: Different stakeholders within the same domain may have specific requirements, needs, and aims, potentially leading to conflicting explanations based on their individual perspectives.
    \item Data-driven vs. Domain-driven Inconsistencies: This type of inconsistency arises when explanations derived from data-driven methods do not align with domain knowledge or stakeholder expectations. 
\end{itemize}

These inconsistencies underscore the complexity of interpreting ML models in scientific contexts. It is often impractical, if not impossible, to satisfy all stakeholders with their diverse aims and needs, or to force users to accept a single model or explanation method as definitive. In the following case studies, we demonstrate how feature-based explanation methods can lead to diverse interpretations of the same scientific phenomena. This exploration highlights the critical importance of considering multiple perspectives when applying ML in scientific domains. Additionally, we identify and discuss common factors contributing to this diversity of explanations, as revealed through our case studies.

\section{Case Studies and Insights}
The following collection of examples highlights some inconsistencies in explanations derived from ML models in the physical sciences, based on four key sources of explanation diversity: 
\begin{itemize}
    \item Model Selection: Accurate ML models trained for the same task can produce diverse explanations at both global and instance levels.
    \item Explanation Method Choice: Well-established XAI methods \footnote{In this study, we use well-established explanation methods, which we define as those that are widely accepted and commonly used in the ML community, such as SHAP, LIME, and Integrated Gradients.} often generate diverse explanations even when applied to the same model.
    \item Feature Attribution Level: Different levels of feature attributions (e.g., first-order vs. higher-order interactions) can result in diverse explanations of the same phenomenon.
    \item Stakeholder Perspective: Diverse needs and priorities of different stakeholders can lead to varied interpretations of model outputs.
\end{itemize}

\subsection{AFLOW Bulk Modulus Benchmark}
\label{sec:aflow}

Our first example explores the prediction of property trends in alloys using their crystal structure, a fundamental problem in materials informatics. A public open-source software in this domain is the AFLOW (Automatic Flow) framework, which enables automated property determination and serves as a benchmark in the field \cite{curtarolo2012aflow, clement2020benchmark}. In a recent study \cite{wang2021compositionally}, the Compositionally Restricted Attention-Based network (CrabNet) was introduced for predicting materials properties structure-agnostic of crystals. This model achieves state-of-the-art performance by using mat2vec embeddings \cite{tshitoyan2019unsupervised} for encoding chemical information and provides heat map style explanations. The authors compared CrabNet with other ML models, including Roost \cite{goodall2020predicting}, ElemNet \cite{jha2018elemnet}, and random forest (RF) models.   

\begin{figure}[h!]
    \centering
     \begin{subfigure}{\textwidth}
        \centering
        \scriptsize
        \begin{tabular}{lllllll}
        \toprule
        Properties   & Roost & CrabNet & HotCrab & ElemNet & RF     & MLP    \\ \midrule
        AFLOW Bulk modulus & 8.82 & 8.69   & 9.10   & 12.12  & 11.91 & 11.30 \\
        \bottomrule
        \end{tabular}
        \caption{}
     \end{subfigure}
     \vfill
     \begin{subfigure}{.48\textwidth}
         \centering
         \includegraphics[width=\textwidth]{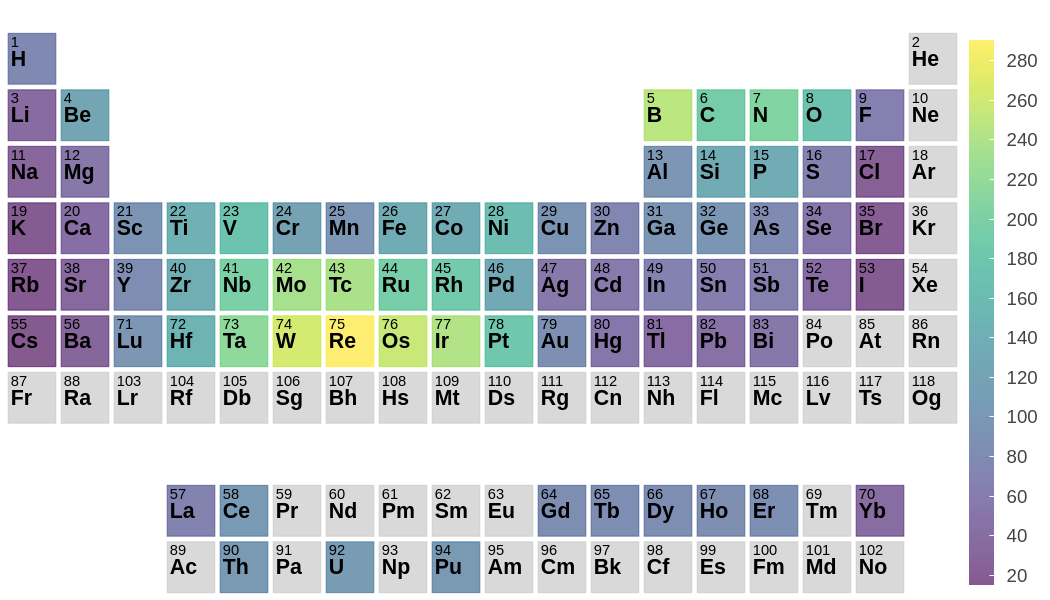}
         \caption{}
     \end{subfigure}
     \hfill
     \begin{subfigure}{.48\textwidth}
         \centering
         \includegraphics[width=\textwidth]{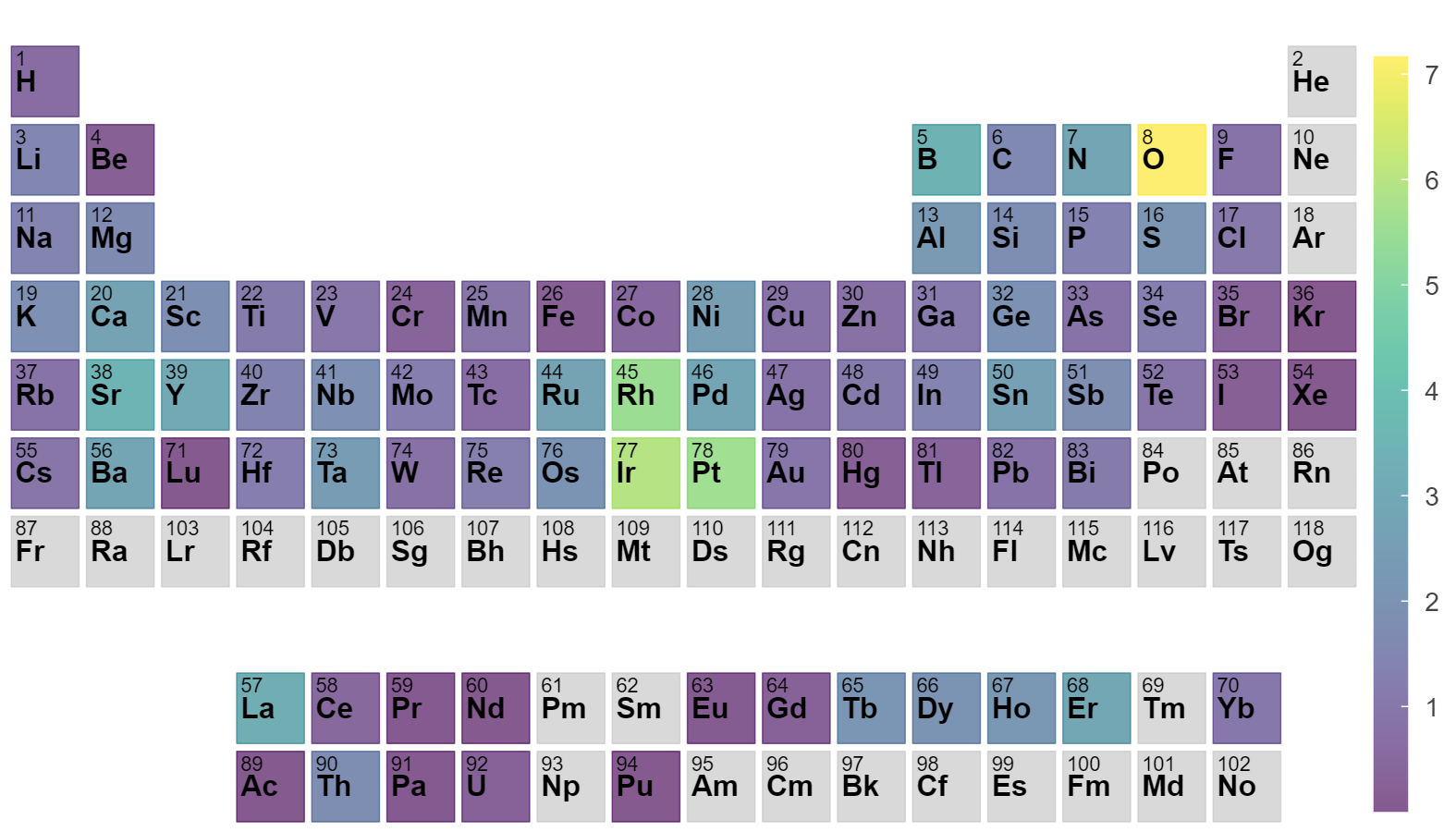}
         \caption{}
     \end{subfigure}
    \caption{
    The average contribution of all elements to bulk modulus predictions, computed from the AFLOW bulk modulus dataset. (a) MAE scores of Roost, CrabNet, one-hot encoded CrabNet (HotCrab), ElemNet, and MLP on the held-out test datasets, compared with the random forest (RF) baseline for the property. (b) Figure reprinted from \cite{wang2021compositionally} under the CC BY 4.0 license \cite{commons2013creative}. (c) First-order feature attribution calculated based on the well-trained MLP. The lighter-coloured elements in the periodic table contribute more towards a compound’s bulk modulus value.} 
\label{fig:aflow-results}
\end{figure}

Here we demonstrate model results evaluated using MAE on the test set in Fig. \ref{fig:aflow-results}(a). 
The models were trained on the same dataset but utilised different encoding representations: mat2vec encoded chemical information (Roost, CrabNet), one-hot encoding without chemical information (one-hot encoded CrabNet (HotCrab), ElemNet, multilayer perception (MLP)), and a Magpie-featurised CBFV \cite{kauwe2018machine} for RF. Here we can see that a simple MLP achieves comparable performance without the need for complex fine-tuning processes. While MLP may offer a slight advantage over RF and ElemNet in performance, we do not anticipate it as well as complex neural networks in benchmark tests. Our focus in this study is on explanations from ML models, particularly the CrabNet and MLP black-boxes.  The self-attention mechanism enables the CrabNet to preserve the elemental identity within a compound and thus directly predict the element’s contribution to the property prediction Fig. \ref{fig:aflow-results}(b) illustrates the average contributions of each element in a CrabNet model, with lighter elements indicating higher contributions. 

Similarly, we compare this to the first-order feature attribution from the MLP model shown in Fig. \ref{fig:aflow-results} (b) and (c), where each element is coloured according to its attribution value. For this case, we split the data set into train, validation, and test sets using a fixed random seed, where the training and validation sets were employed for model training and hyperparameter tuning, respectively, and the test set was used for the evaluation of model performance metrics. We employed Mendeleev encoding \cite{Zhuang2023StructureFreeME} which uses the chemical formula exclusively, and has been shown to achieve comparable performance to other ML models \cite{zhuang2023clf}. Second-order feature interactions in the Supporting Information to highlight the deeper insights into how these interactions contribute to predictions. 

This is significant because different levels of feature attributions lead to diverse explanations. At the first-order level, elements such as iridium (Ir) and platinum (Pt) were found to be independently important, indicating a strong reliance on these elements for accurate predictions. However, the second-order feature interactions (see Supporting Information) show that seemingly less significant elements can play crucial roles when considered in combination. Potassium (K) and chlorine (Cl) have a higher impact on second-order feature attributions despite their lower importance at the first-order level. This diversity in feature attributions across different levels highlights the complexity of the underlying relationships in the data. It also offers researchers the flexibility to select the most appropriate level of feature attribution for their specific research questions or domain knowledge. From a physical perspective, these findings align with our understanding of material properties. The importance of Ir and Pt at the first-order level is consistent with their known high bulk moduli due to strong interatomic bonds. The significance of K and Cl in second-order interactions reflects their role in compound formation and their influence on crystal structures, which can dramatically alter bulk properties \cite{cynn2002osmium, charlie2023effects}.

\subsection{Metallic Nanoparticle Property Prediction}\label{subsec:nano}

\begin{table*}[h!]
\footnotesize
\caption{Representative questions for five four stakeholders in the case of metallic nanoparticles}
\label{tbl:case-study-nano}
\begin{center}
\begin{tabular}{p{2cm}p{2cm}p{2cm}p{3.5cm}p{3cm}}
\toprule
Feature sets    & Representatives     &   Stakeholders     & Representative questions           & Representative \newline expectations                              \\ \midrule
Important & Computer \newline scientist          &Developers& Which features should be included in the model to achieve the best performance?    & I believe that feature importance should depend solely on the model and data.                     \\
Controllable          & Experimental \newline materials \newline scientist &Scientists& Which features are controllable and can be measured using microanalysis methods in the lab?   & I anticipate that controllable features are responsible for functional properties. \\
Structural            & Computational \newline materials \newline scientist    &Scientists& Which features are related to the structure and represent important inputs into my simulations?  & I expect structural features to be related to physical properties.  \\
Experimental         & Manufacturer               &Professionals& Which controllable features are related to the synthesis conditions and attributes that are inputs for industrial processes? & I believe focusing on processing features will save me money. \\ \bottomrule
\end{tabular}
\end{center}
\end{table*}

This case study uses a dataset of metallic nanoparticles originally generated through molecular dynamics simulations that model the sintering and coarsening processes of palladium (Pd), gold (Au), and platinum (Pt) at varying temperatures and atomic deposition rates~\cite{amanda2023pt, amanda2023au, amanda2023pd, li2022impact}. The fractal dimension has recently been proposed as a superior way of describing the complexity of the surfaces that is relevant to catalysis. This dimension is calculated using a box-counting algorithm, using the Sphractal Python package~\cite{ting2024fractal, ting2024sphractal} which is capable of estimating the fractal dimensions for surfaces composed of precise mathematical objects or atomistic coordinates.

In this example, we highlight the impact of using ML models for the same task, including RF and XGBoost, and examine connection weights from a well-trained MLP \cite{beck2018neuralnettools, olden2004accurate} as an exemplar.  One approach to intuitively explain the MLP is by directly analysing the gradients, using IG, given that the model learns through parameter optimisation with respect to the training data. 
To show how this works we considered four domain-based scenarios \textit{Important}, \textit{Controllable}, \textit{Structural}, and \textit{Experimental}, which are described in \cite{li2022impact}. The \textit{Processing} category is not considered because it cannot be fitted well, resulting in unreliable predictions.
Each remaining scenario corresponds to the needs of different potential stakeholders and involves a different feature sub-set. We also include detailed feature sub-sets and descriptions in Supporting Information.

\begin{figure}[t]
     \centering
    \includegraphics[width=\textwidth]{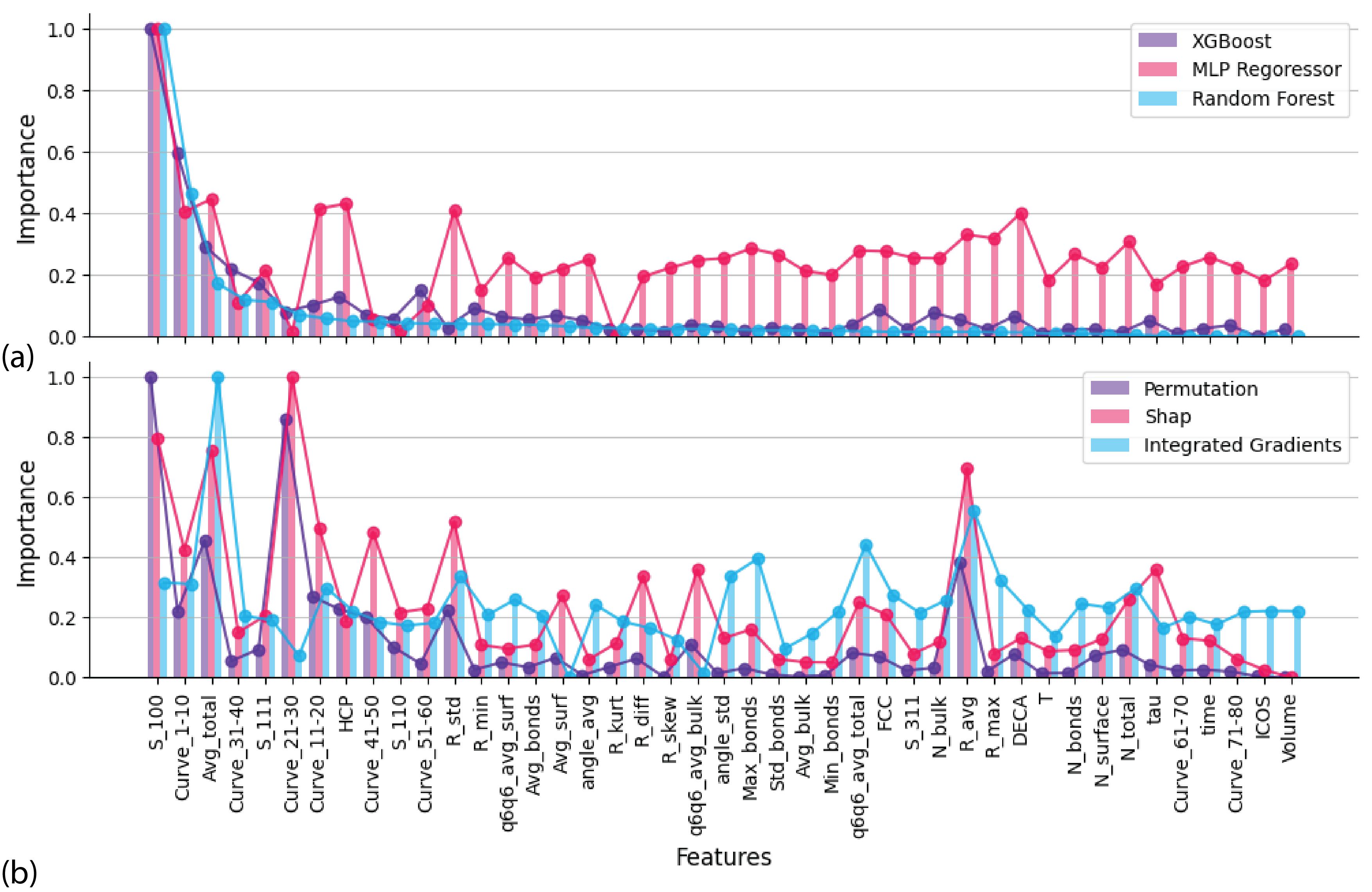}
\caption{\label{fig:nano-model-outcomes} (a) Feature importance rankings of metallic nanoparticles from accurate models, including XGBoost, MLP, and RF, and (b) feature importance rankings (MLP) from different well-established explanation methods including PI, SHAP and IG. The x-axis displays features ordered by their importance ranking from RF, which serves as a baseline. Other rankings are plotted according to this order. The y-axis denotes the importance score, normalised between 0 and 1.}
\end{figure}

\begin{figure}[h!]
     \centering
     \includegraphics[width=\textwidth]{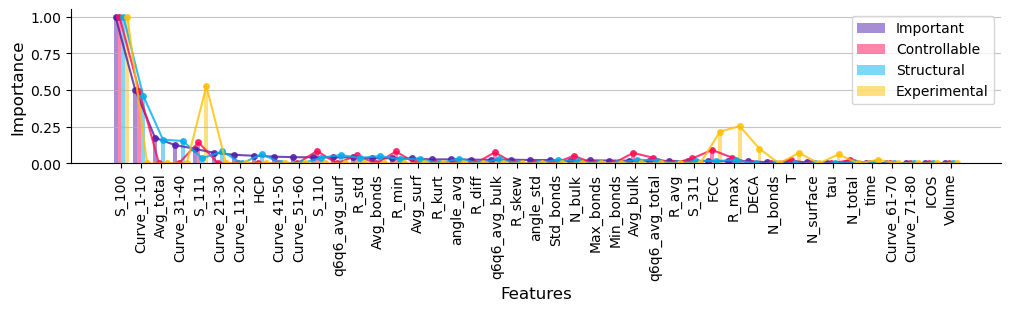}
\caption{\label{fig:nano-model-interaction} Feature importance rankings of metallic nanoparticles fractal dimension predictions from four stakeholders (scenarios). Features ordered by their importance rankings (left to right), serving as a baseline. Other rankings are plotted according to this order. The $y$-axis denotes the importance score, normalised between 0 and 1. 
}
\end{figure}

Using these scenarios, we compare three ML models (XGBoost, MLP and RF) of similar high performance \cite{chicco2021coefficient}.
All models are trained on the same dataset and evaluated using the same metrics (${R^2}$ and MAE), yielding the following results: XGBoost (${R^2}$: 0.74, MAE: 0.03), MLP (${R^2}$: 0.75, MAE: 0.03), and RF (${R^2}$: 0.74, MAE: 0.03). The outcome is illustrated in Fig. \ref{fig:nano-model-outcomes}(top). The normalised importance values from PI and SHAP are also compared, noting the focus here is on the relative rankings rather than the absolute values.  These outcomes are shown in Fig. \ref{fig:nano-model-outcomes}(bottom), and trained four ML models from each scenario separately to predict the formation energy to compare the rankings of features has displayed in Fig. \ref{fig:nano-model-interaction}.

Our example of metallic nanoparticle properties demonstrates that it's possible to identify multiple well-performing models for the same task, leading to diverse explanations. As shown in Fig. \ref{fig:nano-model-outcomes}(a), the feature importance rankings from RF and XGBoost (which are both tree-based) are similar, whereas MLP presents notably different explanations. This variance underscores the challenge in determining which model explanation should be trusted, even when all models perform similarly well on the prediction task, and highlights the need for careful consideration when interpreting ML models in scientific contexts \cite{fisher2019all,li2023exploring,rudin2019stop, li2023exploring}. Relying on a single model's explanation may provide an incomplete or potentially misleading understanding of the underlying phenomena.

In addition to this, Well-established XAI methods lead to diverse explanations from the same model in this example. Based on our well-trained MLP model, example results for IG, SHAP, and PI are shown in Fig. \ref{fig:nano-model-outcomes} (b). We can see here that while all methods agree on the importance of certain features, they differ in their ranking of others. This inconsistency demonstrates that the choice of XAI method can significantly impact the resulting explanation, and highlights the importance of choosing an approach that best aligns with domain knowledge \cite{Barnard2019NanoinformaticsAT}. 


The metallic nanoparticles example also highlights the diversity corresponding to different potential stakeholders with varying needs, aims, or requirements, as summarised in Table \ref{tbl:case-study-nano}. As illustrated in Fig. \ref{fig:nano-model-interaction}, the feature importance rankings differ significantly across these scenarios.  Diverse needs can lead to diverse explanations even when the predictive performance remains constant.  For instance, a manufacturer might focus on features specifically related to synthesis conditions, while model developers might advocate for including features that optimise performance, even if they're beyond laboratory control. A prime example of this divergence is the factor \textit{Time}, which is crucial for laboratory scientists but less significant for computational materials scientists. 
These competing interests impact both model training and the resulting explanation, and emphasise the importance of considering the intended use and audience when developing and interpreting ML models in scientific contexts ~\cite{adadi2018peeking, gilpin2018explaining}. Ultimately scientific researchers are the determining factor in all attempts to explain models, showing consistency with other fields, e.g., a recent perspective in healthcare \cite{miller2023explainable}.

\subsection{MoleculeNet BACE-1 Classification Benchmark}\label{subsec:bace}

The biophysical BACE dataset \cite{subramanian2016computational} comprises 1513 compounds with physicochemical properties used for binary classification focused on inhibitors of human $\beta$-secretase 1 (BACE-1). It includes both quantitative (IC50 values) and qualitative (binary labels) binding results, specifically split into active (IC50 $\leq$ 100 nM) and inactive classes. This dataset integrates experimental values reported in scientific literature, some with detailed crystal structures available. The compounds associated with their 2D structures and binary labels are provided by the benchmark MoleculeNet \cite{wu2018moleculenet}, whihc implements ECFP (Extended-Connectivity Fingerprints) featurisation method to decompose molecules into sub-modules from heavy atoms, with an assigned unique identifier. These identifiers extend through bonds to form larger sub-structures, which are then hashed into fixed-length binary fingerprints which encode the topological characteristics of molecules, enabling applications such as similarity searching and activity prediction. In this example, we use the DeepChem framework for data generation and feature representation~\cite{Ramsundar-et-al-2019}, and present four high-performing RF, XGBoost, Support Vector Machine (SVM), and MLP, based on their performance comparison in MoleculeNet. These models were retrained on the BACE dataset, using the same data for training, and evaluated using accuracy, ROC-AUC, recall, and precision scores.

\begin{table}[]
\centering
\caption{\label{tab:bace}The cross-validation performance results of four models in BACE-1 classification}
\begin{tabular}{lllll} 
\toprule
Models  & Accuracy & ROC-AUC & Recall & Precision \\ \midrule
RF      & 0.813 $\pm$ 0.025     & 0.888 $\pm$ 0.026    & 0.795 $\pm$ 0.029   & 0.801 $\pm$ 0.046    \\
XGBoost & 0.802 $\pm$ 0.024     & 0.876 $\pm$ 0.027    & 0.788 $\pm$ 0.033   & 0.785 $\pm$ 0.043      \\
SVM     & 0.817 $\pm$ 0.029    & 0.884 $\pm$ 0.027    & 0.801 $\pm$ 0.027   & 0.803 $\pm$ 0.045      \\
MLP     & 0.765 $\pm$ 0.061     & 0.862 $\pm$ 0.033    & 0.819 $\pm$ 0.066   & 0.730 $\pm$ 0.084      \\ \bottomrule
\end{tabular}
\end{table}

\subsubsection{Example}

\begin{figure}[ht]
     \centering
     \begin{subfigure}[t]{0.43\textwidth}
         \centering
         \includegraphics[width=\textwidth]{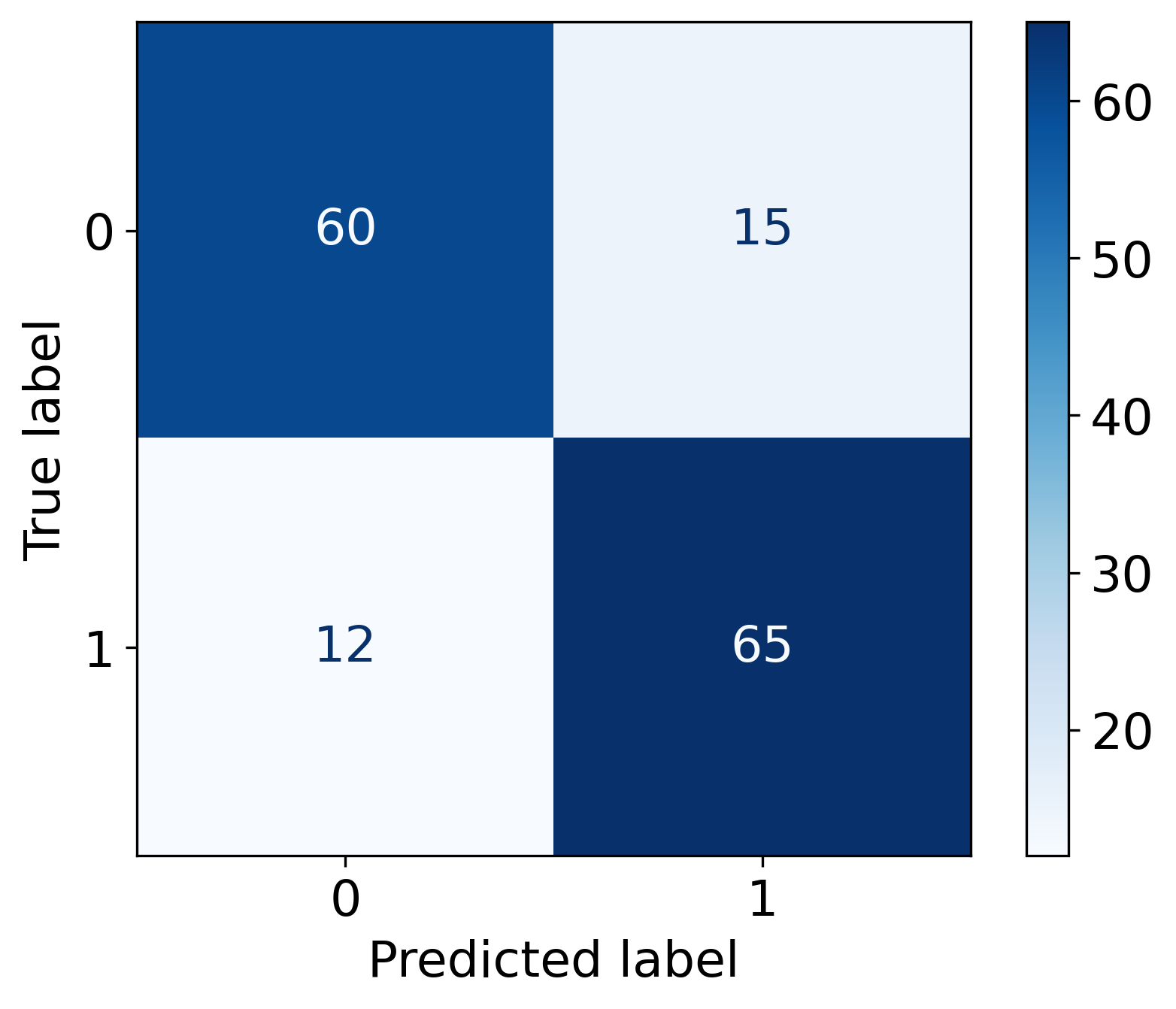}
         \caption{}
     \end{subfigure}
     \hfill
     \begin{subfigure}[t]{0.43\textwidth}
         \centering
         \includegraphics[width=\textwidth]{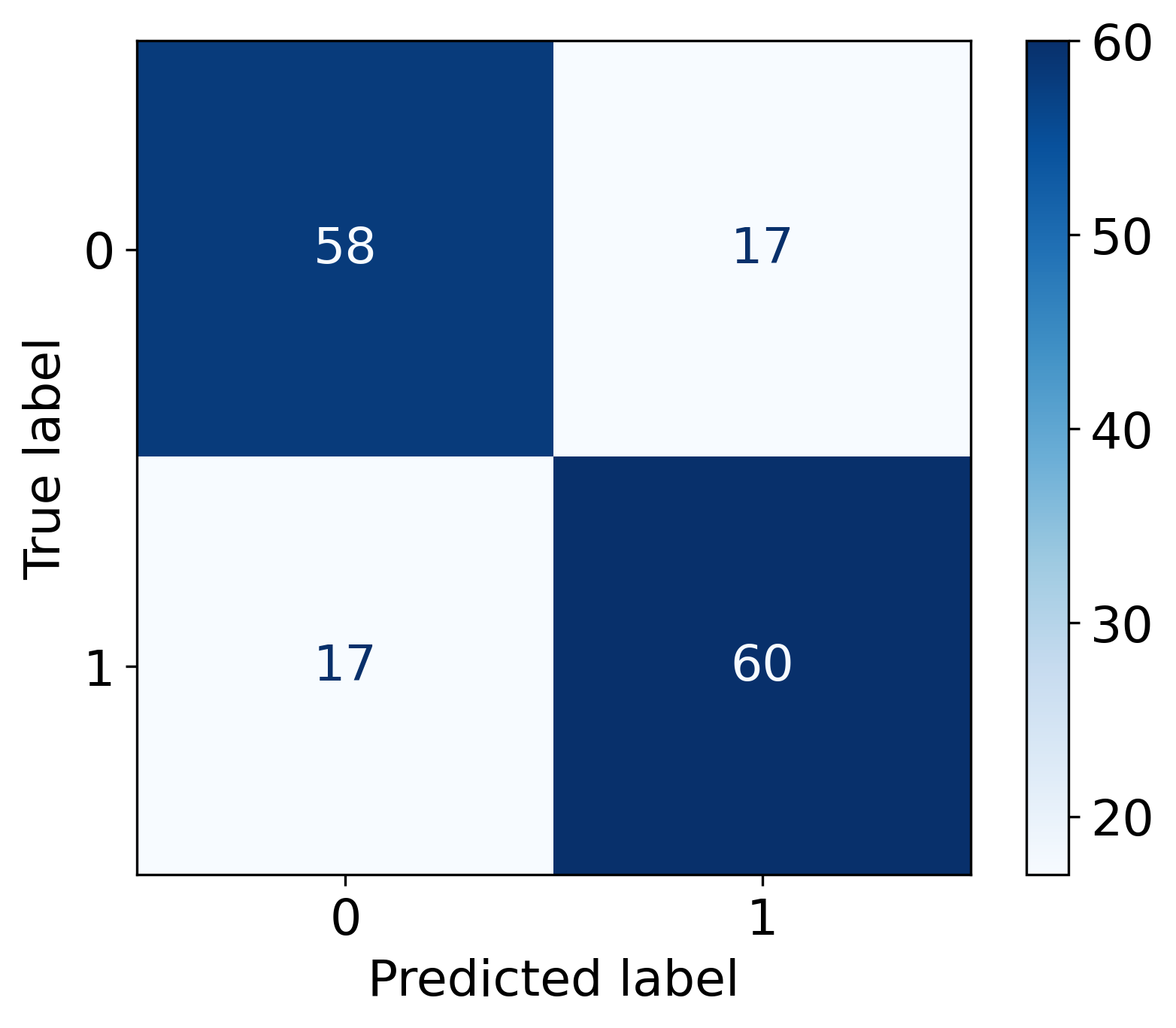}
          \caption{}
     \end{subfigure}
    \\ 
    \begin{subfigure}[t]{0.43\textwidth}
         \centering
         \includegraphics[width=\textwidth]{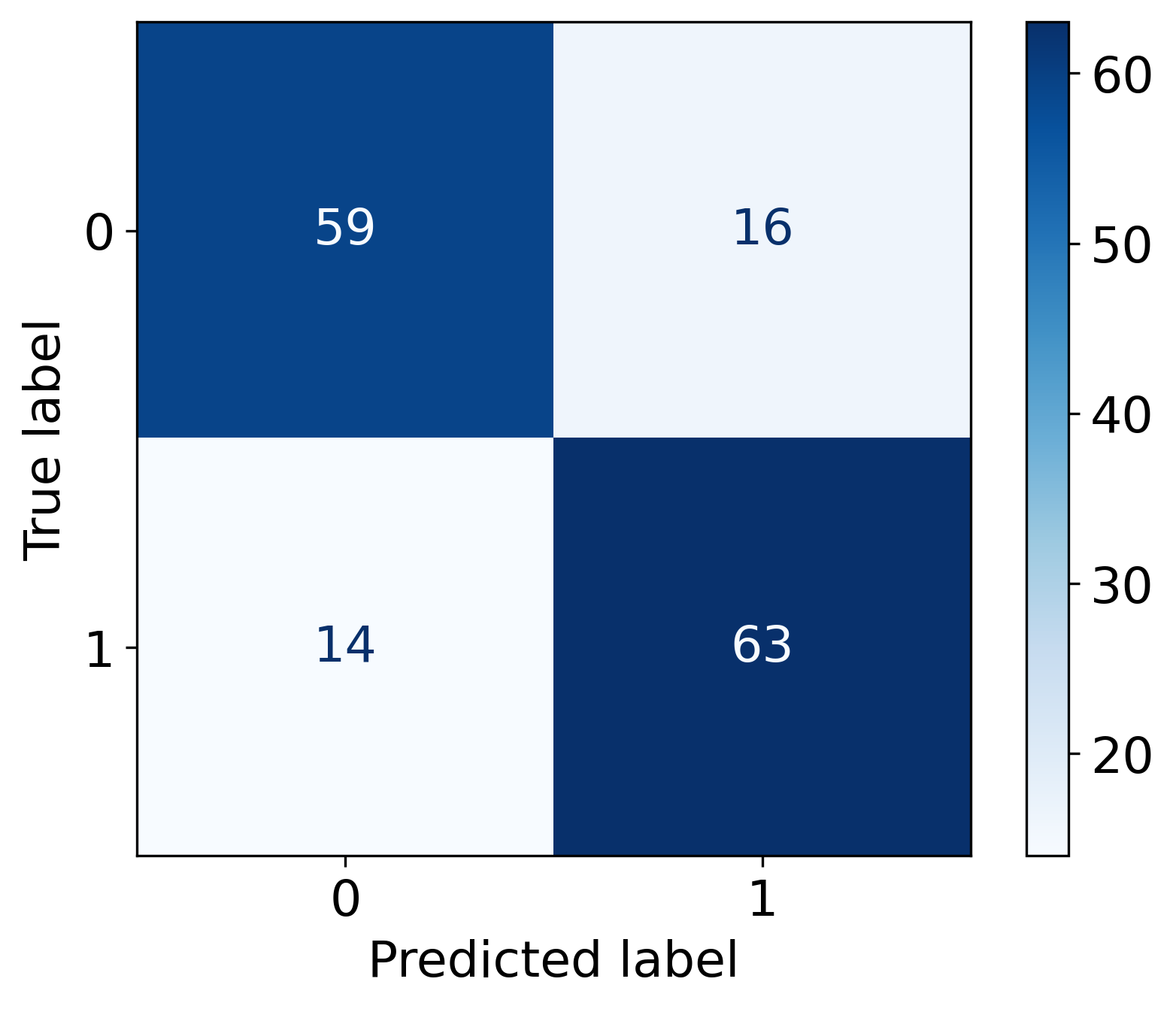}
          \caption{}
     \end{subfigure}
    \hfill
    \begin{subfigure}[t]{0.43\textwidth}
         \centering
         \includegraphics[width=\textwidth]{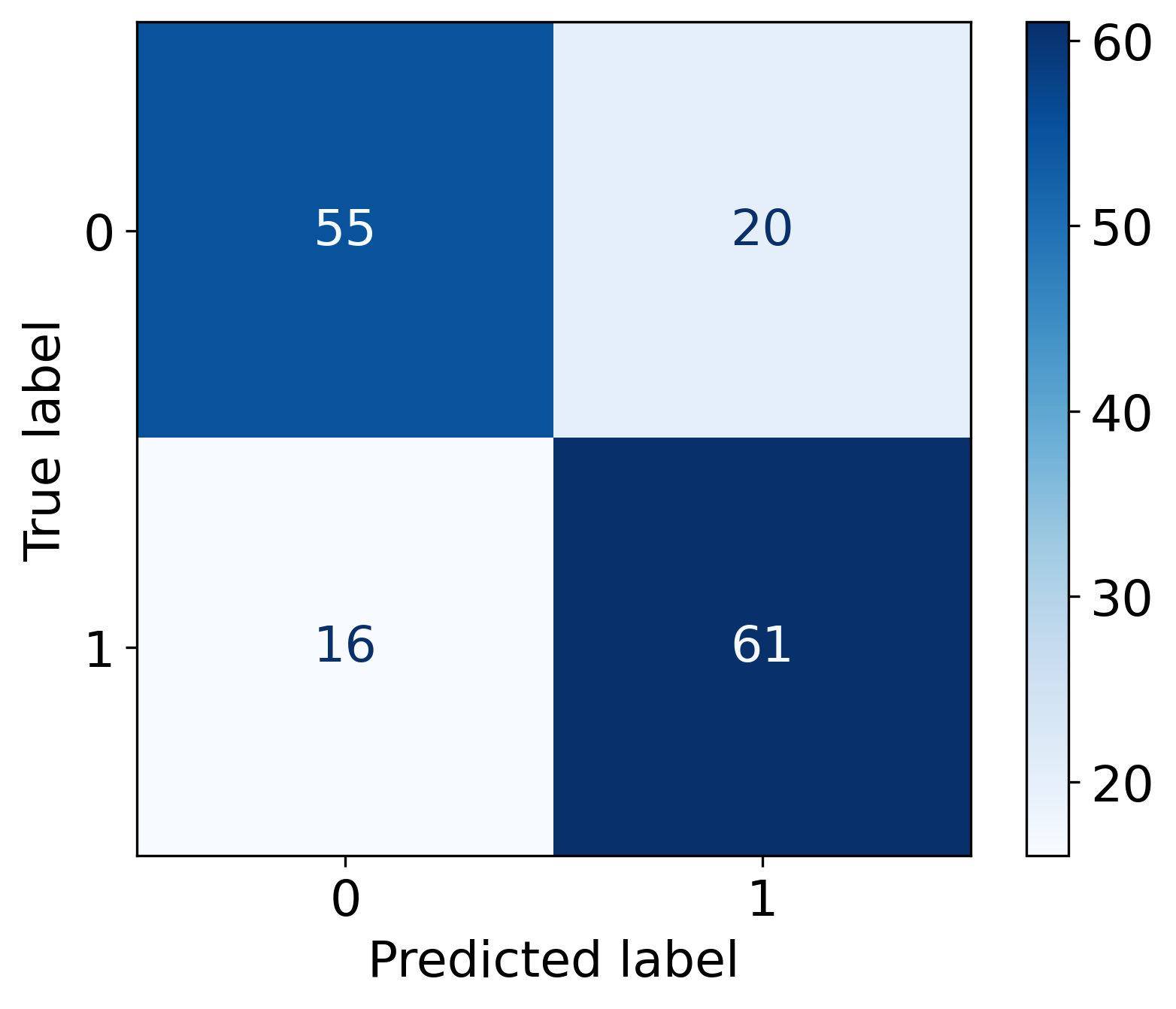}
          \caption{}
     \end{subfigure}
    \caption{
        The confusion matrices of two classifiers in predicting bindings result in a set of inhibitors of human beta-secretase (BACE-1). The models include RF, XGBoost, SVM, and MLP, arranged from (a-d). }
    \label{fig:confusion-matrix}
\end{figure}

\begin{figure}[]
    \centering
    (a)\includegraphics[width=0.4\textwidth]{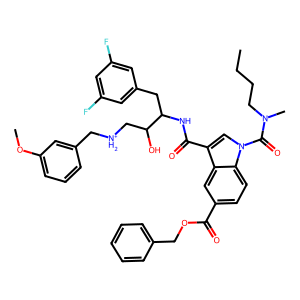}
    (b)\includegraphics[width=0.4\textwidth]{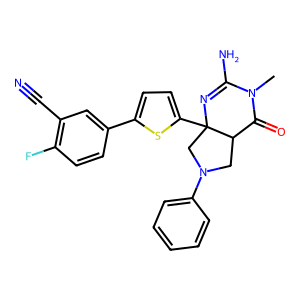}
    \caption{Two samples from BACE dataset misclassified by some of four well-trained models (a) Chemical structure: Fc1cc(cc(F)c1)CC(NC(=O)c1c2cc(ccc2n(c1)C(=O)N(CCCC)C)C(OCc\\1ccccc1)=O)C(O)C[NH2+]Cc1cc(OC)ccc1. The ground truth classification is inactive and is misclassified by RF, SVM, and MLP. (b) Chemical structure: s1c(ccc1-c1cc(C\#N)c(F)cc1)C12N=C(N)N(C)C(=O)C1CN(C2)c1ccccc1.}
    \label{fig:sample-classification}
\end{figure}

\begin{figure}[t!]
     \centering
     \begin{subfigure}[T]{\textwidth}
         \centering
         \includegraphics[width=\textwidth]{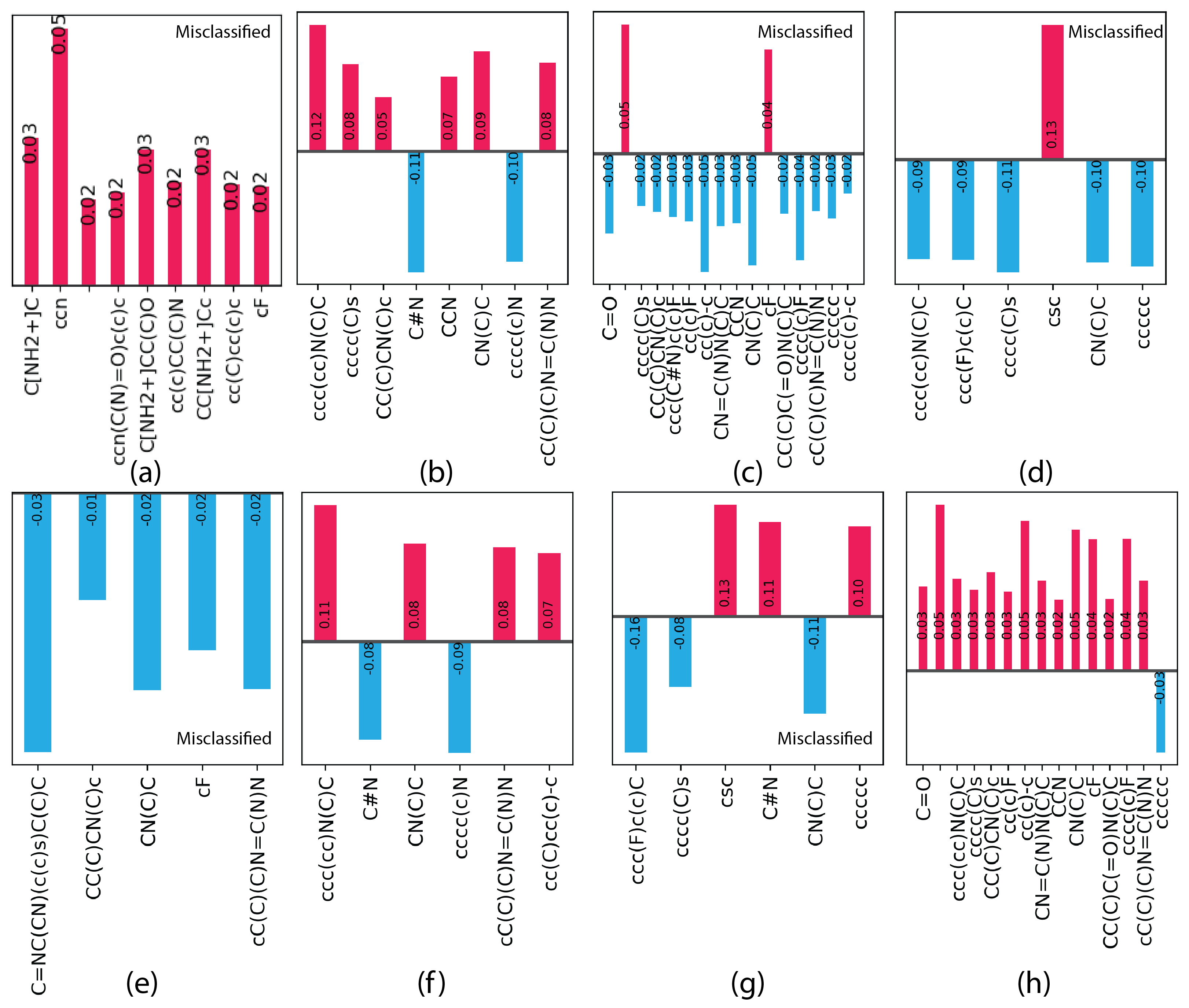}
     \end{subfigure}
    \caption{LIME Local explanations for four well-trained models applied to sample \textit{a} (a-d) and \textit{b} (e-h) from RF, XGBoost, SVM, and MLP model, respectively. Misclassified samples are labelled and it is noted that fragments vary across models due to featurisation. }
    \label{fig:lime-classification}
\end{figure}

The overall performance of all four models are presented in confusion matrices shown in Fig. \ref{fig:confusion-matrix}, where accuracy, ROC-AUC, recall and precision scores are included in Table \ref{tab:bace}. These performances are similar to those reported in MoleculeNet benchmark \cite{wu2018moleculenet}. In this case, our objective is to highlight individual predictions and their associated explanations by applying LIME, which creates locally perturbed datasets and fits linear surrogate models based on these perturbations, shown in \ref{fig:sample-classification}, from the equally well-trained models, as illustrated in Fig. \ref{fig:lime-classification}. In this case,  while the RF and MLP show promising overall accuracy, they misclassified both of the two samples. In contrast, the XGBoost model accurately classified both samples and SVM misclassified one out of the two samples.

Our analysis of the BACE-1 dataset using multiple high-performing models (RF, XGBoost, SVM, and MLP) highlights an important phenomenon known as predictive multiplicity \cite{hsu2022rashomon}. This concept captures the potential for individual-level discrepancies introduced by the arbitrary choice of a single model, even when the overall performance metrics are similar.
As illustrated in Fig. \ref{fig:lime-classification}, we observe significant variations in predictions and explanations across different models for the same molecular samples. For instance, the two samples presented are classified differently by various models, despite all models achieving comparable overall performance metrics (Fig. \ref{fig:confusion-matrix}). This divergence in individual predictions underscores the limitations of relying solely on aggregate performance measures when selecting models for critical scientific applications.
The LIME explanations highlight this issue by revealing how different models prioritise various molecular substructures in their decision-making processes. For example, the fragment cC(C)(C)N=C(N)N in the sample (\textit{b}) contributes positively to the prediction in some models but negatively in others. Such inconsistencies at the local level can have significant implications in domains such as drug discovery, where decisions about individual compounds can have far-reaching consequences. 

This predictive multiplicity in BACE-1 inhibitor classification aligns with the known complexity of protein-ligand interactions, where subtle changes in molecular structure can significantly alter binding affinity \cite{dhanabalan2017identification}. The variation in predictions across models for the same input highlights the inherent uncertainty in ML approaches, especially when applied to complex molecular systems. 
This suggests the need for ensemble approaches, where considering predictions from multiple high-performing models may provide a more robust and nuanced understanding of the underlying relationships in the data. Tools like LIME, which provide instance-level explanations, are crucial for understanding and potentially mitigating the effects of predictive multiplicity. 

\subsection{Insights Into Diverse Explanations}

Our case studies across different domains in physical science (materials science, nanotechnology and chemistry) reveal several key insights into the sources of diverse explanations. Whether predicting bulk modulus, nanoparticle properties, or molecular activities, we consistently observed diverse explanations from data-driven explanations, domain-driven explanations, and data-driven vs. domain-driven explanations. This ubiquity underscores the importance of considering explanation diversity in scientific ML applications.

We have illustrated the four high-level sources of explanation diversity: model selection, explanation method choice, feature attribution level, and stakeholder perspective. These findings align with discoveries in other fields, suggesting a broader applicability of these concepts. 
Each model (e.g., MLP, RF, and XGBoost) and explanation method (e.g., SHAP, LIME, and IG) is theoretically sound but operates under distinct assumptions, which influence feature importance and lead to diverse explanations.
Another factor contributing to this diversity is the complex dependencies between input features \cite{molnar2020limitations, aas2021explaining}, which can significantly impact how different explanation methods attribute importance. This corresponds to the challenges in higher-order feature attributions. Recent research \cite{ghosh2024mapping, feix2017quantum, fox2024active, miklin2017entropic, ghosh2024towards, saha2023cross} has begun exploring causal models in materials informatics to address this limitation, showing promise for extracting deeper materials physics insights.
The existence of multiple well-performing models for the same task aligns with the concept of Rashomon sets  \cite{li2023multi, hsu2022rashomon}, while the variability in explanations from different methods has been noted in other studies \cite{krishna2022disagreement}.

\subsection{Scientists-Centric Perspective}

In light of the inherent diversity in explanations derived from ML, we advocate for a scientists-centric approach to interpretation. The complexity of scientific research often leads to diverse requirements that are challenging to resolve within a single model or explanation method. Empowering scientists to maintain control over the explanation process \cite{9007737, miller2019explanation} includes recognising that different stakeholders in scientific contexts often have varied needs and perspectives when employing ML models.  This perspective discourages blind trust in model-generated explanations, and positions explanations as tools serving scientists, ensuring that insights derived from ML models are relevant and actionable for each specific scientific context or stakeholder. 

The inherent diversity in ML models and explanation methods can be leveraged to achieve consistency across these varied perspectives. A potential pipeline to address this could involve two key steps: 1. initial exploration of a set of well-performing models, effectively sampling the model space to capture a range of interpretations.
2. application of optimization algorithms to identify models that best align with scientists' expectations and domain knowledge.
By embracing this diversity, researchers can address different needs independently, as we saw in the nanoparticle case study. Similarly, in molecular property prediction tasks, as illustrated in our BACE-1 classification example, scientists can utilise multiple high-performing models and local explanation methods like LIME to gain insights into individual predictions. This approach allows for a balance between data-driven insights and domain expertise, producing interpretations that are scientifically sound and practically useful for diverse needs.

\section{Summary and Opportunities}

In this Perspective, we discuss diverse explanations from both data-driven and domain-driven points of view, and the inconsistencies that hinder ML's potential impact in the physical sciences. We illustrate this through three examples drawn from public research: AFLOW for property prediction, metallic nanoparticle property prediction, and BACE-1 classification. We identified four sources contributing to diverse explanations, including different levels of feature attributions, different well-established explanation methods, different well-trained ML models, and different requirements for stakeholders. All of these sources rely on high-performing ML models yet result in different explanations. We advocate for a scientists-centric approach, embracing this diversity as an opportunity to enhance scientific understanding. This approach allows researchers to select methods aligning with their specific needs and domain expertise, maintaining human oversight in the interpretation process \cite{gilpin2018explaining}. 

Looking forward, we suggest that considering sets of models rather than single models offers a promising direction, potentially providing a range of feature attributions without compromising performance. Integrating various explanation methods (such as concept-based, example-based, and feature-based approaches) also shows potential for addressing diverse needs in scientific research.  By fostering an understanding of diverse explanations in scientific ML, we can also contribute to the responsible integration of XAI into scientific discovery, enhancing trustworthiness and deepening our comprehension of complex physical phenomena.


\section*{Data Availability Statement}
This Perspective does not present primary research results or introduce new data, software, or code. To ensure reproducibility we provide access to the following resources:
\begin{itemize}
    \item Supplementary code is available at: \\ \url{https://github.com/Sichao-Li/Diverse-Explanations-Physical-Sciences/}
    \item Detailed information on all public datasets, ML models, and codes referenced in this Perspective is provided in the Supporting Information.
\end{itemize}
These resources allow readers to reproduce our findings and further explore the concepts discussed in this Perspective.

\section*{Acknowledgements}
We gratefully acknowledge the National Computational Infrastructure (NCI) for providing access to their computing facilities in model training under project number p00.

\bibliography{explain} 

\begin{thebibliography}{10}

\bibitem{zhong2022explainable}
Xiaoting Zhong, Brian Gallagher, Shusen Liu, Bhavya Kailkhura, Anna Hiszpanski, and T~Yong-Jin Han.
\newblock Explainable machine learning in materials science.
\newblock {\em npj Computational Materials}, 8(1):204, 2022.

\bibitem{kailkhura2019reliable}
Bhavya Kailkhura, Brian Gallagher, Sookyung Kim, Anna Hiszpanski, and T~Yong-Jin Han.
\newblock Reliable and explainable machine-learning methods for accelerated material discovery.
\newblock {\em npj Computational Materials}, 5(1):108, 2019.

\bibitem{xu2020high}
Yuanfeng Xu, Luis Elcoro, Zhi-Da Song, Benjamin~J Wieder, MG~Vergniory, Nicolas Regnault, Yulin Chen, Claudia Felser, and B~Andrei Bernevig.
\newblock High-throughput calculations of magnetic topological materials.
\newblock {\em Nature}, 586(7831):702--707, 2020.

\bibitem{butler2018machine}
Keith~T Butler, Daniel~W Davies, Hugh Cartwright, Olexandr Isayev, and Aron Walsh.
\newblock Machine learning for molecular and materials science.
\newblock {\em Nature}, 559(7715):547--555, 2018.

\bibitem{schmidt2019recent}
Jonathan Schmidt, M{\'a}rio~RG Marques, Silvana Botti, and Miguel~AL Marques.
\newblock Recent advances and applications of machine learning in solid-state materials science.
\newblock {\em npj computational materials}, 5(1):83, 2019.

\bibitem{li2022inverse}
Sichao Li and Amanda~S Barnard.
\newblock Inverse design of mxenes for high-capacity energy storage materials using multi-target machine learning.
\newblock {\em Chemistry of Materials}, 34(11):4964--4974, 2022.

\bibitem{liu2017materials}
Yue Liu, Tianlu Zhao, Wangwei Ju, and Siqi Shi.
\newblock Materials discovery and design using machine learning.
\newblock {\em Journal of Materiomics}, 3(3):159--177, 2017.

\bibitem{behler2016perspective}
J{\"o}rg Behler.
\newblock Perspective: Machine learning potentials for atomistic simulations.
\newblock {\em The Journal of chemical physics}, 145(17):170901, 2016.

\bibitem{schutt2014represent}
Kristof~T Sch{\"u}tt, Henning Glawe, Felix Brockherde, Antonio Sanna, Klaus-Robert M{\"u}ller, and Eberhard~KU Gross.
\newblock How to represent crystal structures for machine learning: Towards fast prediction of electronic properties.
\newblock {\em Physical Review B}, 89(20):205118, 2014.

\bibitem{yang2019establishing}
Zijiang Yang, Yuksel~C Yabansu, Dipendra Jha, Wei-keng Liao, Alok~N Choudhary, Surya~R Kalidindi, and Ankit Agrawal.
\newblock Establishing structure-property localization linkages for elastic deformation of three-dimensional high contrast composites using deep learning approaches.
\newblock {\em Acta Materialia}, 166:335--345, 2019.

\bibitem{Barnard2020SelectingML}
Amanda~S. Barnard and George Opletal.
\newblock Selecting machine learning models for metallic nanoparticles.
\newblock {\em Nano Futures}, 4:035003, 2020.

\bibitem{Parker2021UnsupervisedSC}
Amanda~J. Parker and Amanda~S. Barnard.
\newblock Unsupervised structure classes vs. supervised property classes of silicon quantum dots using neural networks.
\newblock {\em Nanoscale horizons}, 6:277--282, 2021.

\bibitem{wiens2018machine}
Jenna Wiens and Erica~S Shenoy.
\newblock Machine learning for healthcare: on the verge of a major shift in healthcare epidemiology.
\newblock {\em Clinical infectious diseases}, 66(1):149--153, 2018.

\bibitem{carvalho2022artificial}
Rodrigo~P Carvalho, Cleber~FN Marchiori, Daniel Brandell, and C~Moyses Araujo.
\newblock Artificial intelligence driven in-silico discovery of novel organic lithium-ion battery cathodes.
\newblock {\em Energy storage materials}, 44:313--325, 2022.

\bibitem{varshney2017safety}
Kush~R Varshney and Homa Alemzadeh.
\newblock On the safety of machine learning: {C}yber-physical systems, decision sciences, and data products.
\newblock {\em Big data}, 5(3):246--255, 2017.

\bibitem{jimenez2020drug}
Jos{\'e} Jim{\'e}nez-Luna, Francesca Grisoni, and Gisbert Schneider.
\newblock Drug discovery with explainable artificial intelligence.
\newblock {\em Nature Machine Intelligence}, 2(10):573--584, 2020.

\bibitem{huang2023explainable}
Weitong Huang, Hanna Suominen, Tommy Liu, Gregory Rice, Carlos Salomon, and Amanda~S Barnard.
\newblock Explainable discovery of disease biomarkers: The case of ovarian cancer to illustrate the best practice in machine learning and {S}hapley analysis.
\newblock {\em Journal of Biomedical Informatics}, 141:104365, 2023.

\bibitem{barnard2023importance}
Amanda~S Barnard and Bronwyn~L Fox.
\newblock Importance of {S}tructural {F}eatures and the {I}nfluence of {I}ndividual {S}tructures of {G}raphene {O}xide {U}sing {S}hapley {V}alue {A}nalysis.
\newblock {\em Chemistry of Materials}, 35(21):8840--8856, 2023.

\bibitem{barnard2022explainable}
Amanda~S. Barnard.
\newblock Explainable prediction of {N}-{V}-related defects in nanodiamond using neural networks and {S}hapley values.
\newblock {\em Cell Reports Physical Science}, 3(1):100696, 2022.

\bibitem{rudin2019stop}
Cynthia Rudin.
\newblock Stop explaining black box machine learning models for high stakes decisions and use interpretable models instead.
\newblock {\em Nature machine intelligence}, 1(5):206--215, 2019.

\bibitem{sichao2023vtf}
Sichao Li and Amanda Barnard.
\newblock Variance {T}olerance {F}actors {F}or {I}nterpreting {All} {N}eural {N}etworks.
\newblock In {\em 2023 International Joint Conference on Neural Networks (IJCNN)}, pages 1--9, 2023.

\bibitem{fisher2019all}
Aaron Fisher, Cynthia Rudin, and Francesca Dominici.
\newblock All models are wrong, but many are useful: Learning a variable's importance by studying an entire class of prediction models simultaneously.
\newblock {\em J. Mach. Learn. Res.}, 20(177):1--81, 2019.

\bibitem{li2023exploring}
Sichao Li, Rong Wang, Quanling Deng, and Amanda Barnard.
\newblock Exploring the cloud of feature interaction scores in a {R}ashomon set.
\newblock {\em arXiv preprint arXiv:2305.10181}, 2023.

\bibitem{hsu2022rashomon}
Hsiang Hsu and Flavio Calmon.
\newblock Rashomon {C}apacity: A {M}etric for {P}redictive {M}ultiplicity in {C}lassification.
\newblock {\em Advances in Neural Information Processing Systems}, 35:28988--29000, 2022.

\bibitem{reichstein2019deep}
Markus Reichstein, Gustau Camps-Valls, Bjorn Stevens, Martin Jung, Joachim Denzler, Nuno Carvalhais, and fnm Prabhat.
\newblock Deep learning and process understanding for data-driven {E}arth system science.
\newblock {\em Nature}, 566(7743):195--204, 2019.

\bibitem{roscher2020explainable}
Ribana Roscher, Bastian Bohn, Marco~F Duarte, and Jochen Garcke.
\newblock Explainable machine learning for scientific insights and discoveries.
\newblock {\em Ieee Access}, 8:42200--42216, 2020.

\bibitem{linardatos2020explainable}
Pantelis Linardatos, Vasilis Papastefanopoulos, and Sotiris Kotsiantis.
\newblock Explainable ai: A review of machine learning interpretability methods.
\newblock {\em Entropy}, 23(1):18, 2020.

\bibitem{lipton2018mythos}
Zachary~C Lipton.
\newblock The mythos of model interpretability: In machine learning, the concept of interpretability is both important and slippery.
\newblock {\em Queue}, 16(3):31--57, 2018.

\bibitem{kim2016examples}
Been Kim, Rajiv Khanna, and Oluwasanmi~O Koyejo.
\newblock In {\em Examples are not enough, learn to criticize! criticism for interpretability}, volume~29, 2016.

\bibitem{miller2019explanation}
Tim Miller.
\newblock Explanation in artificial intelligence: Insights from the social sciences.
\newblock {\em Artificial intelligence}, 267:1--38, 2019.

\bibitem{imrie2023multiple}
Fergus Imrie, Robert Davis, and Mihaela van~der Schaar.
\newblock Multiple stakeholders drive diverse interpretability requirements for machine learning in healthcare.
\newblock {\em Nature Machine Intelligence}, 5(8):824--829, 2023.

\bibitem{adadi2018peeking}
Amina Adadi and Mohammed Berrada.
\newblock Peeking {I}nside the {B}lack-{B}ox: {A} {S}urvey on {E}xplainable {A}rtificial {I}ntelligence ({XAI}).
\newblock {\em IEEE Access}, 6:52138--52160, 2018.

\bibitem{doshi2017towards}
Finale Doshi-Velez and Been Kim.
\newblock Towards a rigorous science of interpretable machine learning.
\newblock {\em arXiv preprint arXiv:1702.08608}, 2017.

\bibitem{Liu2023TheER}
Tommy Liu and Amanda~S. Barnard.
\newblock The emergent role of explainable artificial intelligence in the materials sciences.
\newblock {\em Cell Reports Physical Science}, 4:101630, 2023.

\bibitem{freitas2014comprehensible}
Alex~A Freitas.
\newblock Comprehensible classification models: a position paper.
\newblock {\em ACM SIGKDD explorations newsletter}, 15(1):1--10, 2014.

\bibitem{gilpin2018explaining}
Leilani~H Gilpin, David Bau, Ben~Z Yuan, Ayesha Bajwa, Michael Specter, and Lalana Kagal.
\newblock Explaining explanations: An overview of interpretability of machine learning.
\newblock In {\em 2018 IEEE 5th International Conference on data science and advanced analytics (DSAA)}, pages 80--89. IEEE, 2018.

\bibitem{guidotti2018survey}
Riccardo Guidotti, Anna Monreale, Salvatore Ruggieri, Franco Turini, Fosca Giannotti, and Dino Pedreschi.
\newblock A survey of methods for explaining black box models.
\newblock {\em ACM computing surveys (CSUR)}, 51(5):1--42, 2018.

\bibitem{nauta2023anecdotal}
Meike Nauta, Jan Trienes, Shreyasi Pathak, Elisa Nguyen, Michelle Peters, Yasmin Schmitt, J{\"o}rg Schl{\"o}tterer, Maurice van Keulen, and Christin Seifert.
\newblock From {A}necdotal {E}vidence to {Q}uantitative {E}valuation {M}ethods: {A} {S}ystematic {R}eview on {E}valuating {E}xplainable {AI}.
\newblock {\em ACM Computing Surveys}, 55(13s):1--42, 2023.

\bibitem{gola2018advanced}
Jessica Gola, Dominik Britz, Thorsten Staudt, Marc Winter, Andreas~Simon Schneider, Marc Ludovici, and Frank M{\"u}cklich.
\newblock Advanced microstructure classification by data mining methods.
\newblock {\em Computational Materials Science}, 148:324--335, 2018.

\bibitem{pankajakshan2017machine}
Praveen Pankajakshan, Suchismita Sanyal, Onno~E de~Noord, Indranil Bhattacharya, Arnab Bhattacharyya, and Umesh Waghmare.
\newblock Machine learning and statistical analysis for materials science: stability and transferability of fingerprint descriptors and chemical insights.
\newblock {\em Chemistry of Materials}, 29(10):4190--4201, 2017.

\bibitem{9007737}
Ribana Roscher, Bastian Bohn, Marco~F. Duarte, and Jochen Garcke.
\newblock Explainable {M}achine {L}earning for {S}cientific {I}nsights and {D}iscoveries.
\newblock {\em IEEE Access}, 8:42200--42216, 2020.

\bibitem{wang2021compositionally}
Anthony Yu-Tung Wang, Steven~K Kauwe, Ryan~J Murdock, and Taylor~D Sparks.
\newblock Compositionally restricted attention-based network for materials property predictions.
\newblock {\em Npj Computational Materials}, 7(1):77, 2021.

\bibitem{karpatne2017theory}
Anuj Karpatne, Gowtham Atluri, James~H Faghmous, Michael Steinbach, Arindam Banerjee, Auroop Ganguly, Shashi Shekhar, Nagiza Samatova, and Vipin Kumar.
\newblock Theory-guided data science: A new paradigm for scientific discovery from data.
\newblock {\em IEEE Transactions on knowledge and data engineering}, 29(10):2318--2331, 2017.

\bibitem{Zhang2011AvoidingSC}
Jia-Zhong Zhang.
\newblock Avoiding spurious correlation in analysis of chemical kinetic data.
\newblock {\em Chemical communications}, 47 24:6861--6863, 2011.

\bibitem{breiman2017classification}
Leo Breiman.
\newblock {\em Classification and regression trees}.
\newblock Routledge, 2017.

\bibitem{chen2015xgboost}
Tianqi Chen, Tong He, Michael Benesty, Vadim Khotilovich, Yuan Tang, Hyunsu Cho, Kailong Chen, Rory Mitchell, Ignacio Cano, Tianyi Zhou, et~al.
\newblock Xgboost: extreme gradient boosting.
\newblock {\em R package version 0.4-2}, 1(4):1--4, 2015.

\bibitem{lundberg2017unified}
Scott~M Lundberg and Su-In Lee.
\newblock In {\em A unified approach to interpreting model predictions}, volume~30, 2017.

\bibitem{breiman2001random}
Leo Breiman.
\newblock Random forests.
\newblock {\em Machine learning}, 45:5--32, 2001.

\bibitem{ribeiro2016should}
Marco~Tulio Ribeiro, Sameer Singh, and Carlos Guestrin.
\newblock " why should i trust you?" {E}xplaining the predictions of any classifier.
\newblock In {\em Proceedings of the 22nd ACM SIGKDD international conference on knowledge discovery and data mining}, pages 1135--1144, 2016.

\bibitem{sundararajan2017axiomatic}
Mukund Sundararajan, Ankur Taly, and Qiqi Yan.
\newblock Axiomatic {A}ttribution for {D}eep {N}etworks.
\newblock In Doina Precup and Yee~Whye Teh, editors, {\em Proceedings of the 34th International Conference on Machine Learning}, volume~70 of {\em Proceedings of Machine Learning Research}, pages 3319--3328. PMLR, 06--11 Aug 2017.

\bibitem{beck2018neuralnettools}
Marcus~W Beck.
\newblock {NeuralNetTools}: {V}isualization and analysis tools for neural networks.
\newblock {\em Journal of statistical software}, 85(11):1, 2018.

\bibitem{olden2004accurate}
Julian~D Olden, Michael~K Joy, and Russell~G Death.
\newblock An accurate comparison of methods for quantifying variable importance in artificial neural networks using simulated data.
\newblock {\em Ecological modelling}, 178(3-4):389--397, 2004.

\bibitem{ribeiro2016model}
Marco~Tulio Ribeiro, Sameer Singh, and Carlos Guestrin.
\newblock Model-agnostic interpretability of machine learning.
\newblock {\em arXiv preprint arXiv:1606.05386}, 2016.

\bibitem{janizek2021explaining}
Joseph~D Janizek, Pascal Sturmfels, and Su-In Lee.
\newblock Explaining explanations: Axiomatic feature interactions for deep networks.
\newblock {\em The Journal of Machine Learning Research}, 22(1):4687--4740, 2021.

\bibitem{li2024practical}
Sichao Li, Amanda~S Barnard, and Quanling Deng.
\newblock Practical attribution guidance for rashomon sets.
\newblock {\em arXiv preprint arXiv:2407.18482}, 2024.

\bibitem{dong2020exploring}
Jiayun Dong and Cynthia Rudin.
\newblock Exploring the cloud of variable importance for the set of all good models.
\newblock {\em Nature Machine Intelligence}, 2(12):810--824, 2020.

\bibitem{datta2016algorithmic}
Anupam Datta, Shayak Sen, and Yair Zick.
\newblock Algorithmic transparency via quantitative input influence: Theory and experiments with learning systems.
\newblock In {\em 2016 IEEE symposium on security and privacy (SP)}, pages 598--617. IEEE, 2016.

\bibitem{curtarolo2012aflow}
Stefano Curtarolo, Wahyu Setyawan, Gus~LW Hart, Michal Jahnatek, Roman~V Chepulskii, Richard~H Taylor, Shidong Wang, Junkai Xue, Kesong Yang, Ohad Levy, et~al.
\newblock Aflow: An automatic framework for high-throughput materials discovery.
\newblock {\em Computational Materials Science}, 58:218--226, 2012.

\bibitem{clement2020benchmark}
Conrad~L Clement, Steven~K Kauwe, and Taylor~D Sparks.
\newblock Benchmark aflow data sets for machine learning.
\newblock {\em Integrating Materials and Manufacturing Innovation}, 9(2):153--156, 2020.

\bibitem{tshitoyan2019unsupervised}
Vahe Tshitoyan, John Dagdelen, Leigh Weston, Alexander Dunn, Ziqin Rong, Olga Kononova, Kristin~A Persson, Gerbrand Ceder, and Anubhav Jain.
\newblock Unsupervised word embeddings capture latent knowledge from materials science literature.
\newblock {\em Nature}, 571(7763):95--98, 2019.

\bibitem{goodall2020predicting}
Rhys~EA Goodall and Alpha~A Lee.
\newblock Predicting materials properties without crystal structure: Deep representation learning from stoichiometry.
\newblock {\em Nature communications}, 11(1):6280, 2020.

\bibitem{jha2018elemnet}
Dipendra Jha, Logan Ward, Arindam Paul, Wei-keng Liao, Alok Choudhary, Chris Wolverton, and Ankit Agrawal.
\newblock Elemnet: Deep learning the chemistry of materials from only elemental composition.
\newblock {\em Scientific reports}, 8(1):17593, 2018.

\bibitem{commons2013creative}
Creative Commons.
\newblock Creative commons attribution 4.0 international license, 2013.

\bibitem{kauwe2018machine}
Steven~K Kauwe, Jake Graser, Antonio Vazquez, and Taylor~D Sparks.
\newblock Machine learning prediction of heat capacity for solid inorganics.
\newblock {\em Integrating Materials and Manufacturing Innovation}, 7:43--51, 2018.

\bibitem{Zhuang2023StructureFreeME}
Zixin Zhuang and Amanda~S. Barnard.
\newblock Structure-free mendeleev encodings of material compounds for machine learning.
\newblock {\em Chemistry of Materials}, 35:9325–9338, 2023.

\bibitem{zhuang2023clf}
Zixin Zhuang and Amanda~S. Barnard.
\newblock Classification of battery compounds using structure-free mendeleev encodings.
\newblock {\em J. Cheminform.}, 16:47, 2024.

\bibitem{cynn2002osmium}
Hyunchae Cynn, John~E Klepeis, Choong-Shik Yoo, and David~A Young.
\newblock Osmium has the lowest experimentally determined compressibility.
\newblock {\em Physical review letters}, 88(13):135701, 2002.

\bibitem{charlie2023effects}
Destiny~E Charlie, Hitler Louis, Goodness~J Ogunwale, Ismail~O Amodu, Providence~B Ashishie, Ernest~C Agwamba, and Adedapo~S Adeyinka.
\newblock Effects of alkali-metals (x= li, na, k) doping on the electronic, optoelectronic, thermodynamic, and x-ray spectroscopic properties of x--sni3 halide perovskites.
\newblock {\em Computational Condensed Matter}, 35:e00798, 2023.

\bibitem{amanda2023pt}
Amanda Barnard, Baichuan Sun, and George Opletal.
\newblock Platinum {N}anoparticle {D}ata {S}et. v2.
\newblock CSIRO. Data Collection., 2018.
\newblock {doi}: 10.25919/5d3958d9bf5f7.

\bibitem{amanda2023au}
Amanda Barnard and George Opletal.
\newblock Gold {N}anoparticle {D}ata {S}et. v1.
\newblock CSIRO. Data Collection., 2019.
\newblock {doi}: 10.25919/5d395ef9a4291.

\bibitem{amanda2023pd}
Amanda Barnard and George Opletal.
\newblock Palladium {N}anoparticle {D}ata {S}et. v2.
\newblock CSIRO. Data Collection., 2023.
\newblock {doi}: 10.25919/epxd-8p61.

\bibitem{li2022impact}
Sichao Li, Jonathan~YC Ting, and Amanda~S Barnard.
\newblock The impact of domain-driven and data-driven feature selection on the inverse design of nanoparticle catalysts.
\newblock {\em Journal of Computational Science}, 65:101896, 2022.

\bibitem{ting2024fractal}
Jonathan Yik~Chang Ting, George Opletal, and Amanda~S. Barnard.
\newblock Fractal characterisation of simulated metal nanocatalysts in 3d.
\newblock {\em Small Science}, 2022.

\bibitem{ting2024sphractal}
Jonathan Yik~Chang Ting, Andrew Thomas~Agars Wood, and Amanda~Susan Barnard.
\newblock Sphractal: {E}stimating the {F}ractal {D}imension of {S}urfaces {C}omputed from {P}recise {A}tomic {C}oordinates via {B}ox-{C}ounting {A}lgorithm, 2024.

\bibitem{chicco2021coefficient}
Davide Chicco, Matthijs~J Warrens, and Giuseppe Jurman.
\newblock The coefficient of determination {R}-squared is more informative than {SMAPE}, {MAE}, {MAPE}, {MSE} and {RMSE} in regression analysis evaluation.
\newblock {\em PeerJ Computer Science}, 7:e623, 2021.

\bibitem{Barnard2019NanoinformaticsAT}
Amanda~S. Barnard, Benyamin Motevalli, Amanda~J Parker, Meli Fischer, Chris Feigl, and George Opletal.
\newblock Nanoinformatics, and the big challenges for the science of small things.
\newblock {\em Nanoscale}, 11:19190--19201, 2019.

\bibitem{miller2023explainable}
Tim Miller.
\newblock Explainable {AI} is {D}ead, {L}ong {L}ive {E}xplainable {AI}! {H}ypothesis-driven {D}ecision {S}upport using {E}valuative {AI}.
\newblock In {\em Proceedings of the 2023 ACM Conference on Fairness, Accountability, and Transparency}, pages 333--342, 2023.

\bibitem{subramanian2016computational}
Govindan Subramanian, Bharath Ramsundar, Vijay Pande, and Rajiah~Aldrin Denny.
\newblock Computational modeling of $\beta$-secretase 1 (bace-1) inhibitors using ligand based approaches.
\newblock {\em Journal of chemical information and modeling}, 56(10):1936--1949, 2016.

\bibitem{wu2018moleculenet}
Zhenqin Wu, Bharath Ramsundar, Evan~N Feinberg, Joseph Gomes, Caleb Geniesse, Aneesh~S Pappu, Karl Leswing, and Vijay Pande.
\newblock Moleculenet: a benchmark for molecular machine learning.
\newblock {\em Chemical science}, 9(2):513--530, 2018.

\bibitem{Ramsundar-et-al-2019}
Bharath Ramsundar, Peter Eastman, Patrick Walters, Vijay Pande, Karl Leswing, and Zhenqin Wu.
\newblock {\em Deep Learning for the Life Sciences}.
\newblock O'Reilly Media, 2019.
\newblock \url{https://www.amazon.com/Deep-Learning-Life-Sciences-Microscopy/dp/1492039837}.

\bibitem{dhanabalan2017identification}
Anantha~Krishnan Dhanabalan, Manish Kesherwani, Devadasan Velmurugan, and Krishnasamy Gunasekaran.
\newblock Identification of new bace1 inhibitors using pharmacophore and molecular dynamics simulations approach.
\newblock {\em Journal of Molecular Graphics and Modelling}, 76:56--69, 2017.

\bibitem{molnar2020limitations}
Christoph Molnar, S~Gruber, and P~Kopper.
\newblock Limitations of interpretable machine learning methods, 2020.

\bibitem{aas2021explaining}
Kjersti Aas, Martin Jullum, and Anders L{\o}land.
\newblock Explaining individual predictions when features are dependent: More accurate approximations to shapley values.
\newblock {\em Artificial Intelligence}, 298:103502, 2021.

\bibitem{ghosh2024mapping}
Ayana Ghosh and Saurabh Ghosh.
\newblock Mapping causal pathways with structural modes fingerprint for perovskite oxides.
\newblock {\em Machine Learning: Science and Technology}, 2024.

\bibitem{feix2017quantum}
Adrien Feix and {\v{C}}aslav Brukner.
\newblock Quantum superpositions of common-causeand direct-cause causal structures.
\newblock {\em New Journal of Physics}, 19(12):123028, 2017.

\bibitem{fox2024active}
Zachary~R Fox and Ayana Ghosh.
\newblock Active causal learning for decoding chemical complexities with targeted interventions.
\newblock {\em arXiv preprint arXiv:2404.04224}, 2024.

\bibitem{miklin2017entropic}
Nikolai Miklin, Alastair~A Abbott, Cyril Branciard, Rafael Chaves, and Costantino Budroni.
\newblock The entropic approach to causal correlations.
\newblock {\em New Journal of Physics}, 19(11):113041, 2017.

\bibitem{ghosh2024towards}
Ayana Ghosh.
\newblock Towards physics-informed explainable machine learning and causal models for materials research.
\newblock {\em Computational Materials Science}, 233:112740, 2024.

\bibitem{saha2023cross}
Anik Saha, Oktie Hassanzadeh, Alex Gittens, Jian Ni, Kavitha Srinivas, and Bulent Yener.
\newblock A cross-domain evaluation of approaches for causal knowledge extraction.
\newblock {\em arXiv preprint arXiv:2308.03891}, 2023.

\bibitem{li2023multi}
Sichao Li and Amanda~S Barnard.
\newblock Multi-target neural network predictions of {MX}enes as high-capacity energy storage materials in a {R}ashomon set.
\newblock {\em Cell Reports Physical Science}, 4(11):101675, 2023.

\bibitem{krishna2022disagreement}
Satyapriya Krishna, Tessa Han, Alex Gu, Javin Pombra, Shahin Jabbari, Steven Wu, and Himabindu Lakkaraju.
\newblock The disagreement problem in explainable machine learning: A practitioner's perspective.
\newblock {\em arXiv preprint arXiv:2202.01602}, 2022.

\end{thebibliography}
\bibliographystyle{unsrt}

\end{document}


\title{SUPPORTING INFORMATION \\Diverse Explanations From Data-Driven and Domain-Driven Perspectives in the Physical Sciences}
\author[1]{Sichao.li\footnote{sichao.li@anu.edu.au}}
\author[1]{Xin Wang}
\author[1]{Amanda S. Barnard}
\affil[1]{School of Computing, Australian National University, Acton 2601, Australia}
\date{}
\setcounter{Maxaffil}{0}
\renewcommand\Affilfont{\itshape\small}
\maketitle

As this is a Perspective article, containing no primary research results, data, software or code, this document contains information about the case studies that have been reproduced.  We include details of dataset information, structures of machine learning models with parameters, feature sets defined for stakeholders, and feature descriptions for datasets.  These have been included for demonstration purposes.

\section*{Machine Learning Models and Public Datasets}
The code can be found at \href{https://github.com/Sichao-Li/Diverse-Explanations-Physical-Sciences/tree/master}{the project page}.

\subsection{Datasets}
All datasets used in this study are public, including:
\begin{itemize}
    \item AFLOW Bulk Modulus dataset \cite{curtarolo2012aflow, clement2020benchmark}
    \item Metallic Nanoparticle dataset \cite{amanda2023au, amanda2023pd, amanda2023pt}
    \item BACE dataset \cite{subramanian2016computational}
\end{itemize}
These datasets can also be downloaded on our project page.

\subsection{Machine Learning Models}
Models used in AFLOW Bulk Modulus Benchmark:
\begin{itemize}
    \item MLP: hidden\_layer\_sizes=(128, 128, 128), max\_iter=2000, activation=tanh, learning\_rate\_init=0.005, random\_state=0, early\_stopping=True, solver=adam, batch\_size=16, learning\_rate='constant'
    \item Compositionally Restricted Attention-Based network (CrabNet) Roost, ElemNet, and random forest (RF) models are from \cite{wang2021compositionally}.
\end{itemize}

\noindent Models used in Metallic Nanoparticle property prediction:
\begin{itemize}
    \item RF: max\_depth=30, max\_features=30, min\_samples\_leaf=5, min\_samples\_split=5, n\_estimators=350\
    \item XGBoost: objective=reg:squarederror, random\_state=42, learning\_rate=0.02, max\_depth=5, n\_estimators=350, gamma=0, colsample\_bytree=0.8
    \item MLP: hidden\_layer\_sizes=(64, 64, 64, 64), max\_iter=1000, activation=relu, learning\_rate\_init=0.005, random\_state=42, early\_stopping=True, solver=adam, alpha=0.0001, learning\_rate='constant'
\end{itemize}

\noindent Models used in BACE-1 classification:
\begin{itemize}
    \item MLP: hidden\_layer\_sizes=(32,32,32), alpha=0.01, early\_stopping=True, max\_iter=1000, random\_state=42, learning\_rate\_init=0.0005, solver=adam, learning\_rate=constant
    \item RF: random\_state=42, n\_estimators =100, criterion=entropy
    \item SVM: kernel=rbf, degree=3, gamma=scale, shrinking=True, tol=1e-3, max\_iter=-1, decision\_function\_shape =ovr
    \item XGBoost: max\_depth=3,
    learning\_rate=0.1,
    n\_estimators=100,
    objective=binary:logistic,
    booster=gbtree,
    importance\_type=gain
\end{itemize}

\section*{Feature Sets Defined For Stakeholders}
In the case study of Metallic Nanoparticle property prediction, we considered the following four scenarios ass different stakeholders' needs:
\begin{itemize}
    \item \textit{Important}: S\_100, Curve\_1-10, Avg\_total, Curve\_31-40, S\_111, Curve\_21-30, Curve\_11-20, HCP, Curve\_41-50, R\_std,
       R\_min, q6q6\_avg\_surf, S\_110, Avg\_bonds, Curve\_51-60,
       Avg\_surf, R\_kurt, angle\_avg, R\_diff, R\_skew, angle\_std,
       q6q6\_avg\_bulk, Std\_bonds, Max\_bonds, Avg\_bulk, Min\_bonds,
       q6q6\_avg\_total, R\_avg, FCC, S\_311, N\_bulk, R\_max,
       DECA, N\_bonds, T, N\_surface, N\_total, 'tau,
       Curve\_61-70, time, ICOS, Curve\_71-80
    \item \textit{Controllable}: N\_total, N\_bulk, Curve\_1-10, R\_min, R\_max, S\_111, FCC, T, Avg\_bulk, R\_std, q6q6\_avg\_bulk, S\_110, S\_10, S\_311, q6q6\_avg\_total
    
    \item \textit{Structural}: N\_total, N\_bulk, Curve\_1-10, Avg\_total, R\_min, R\_max, Curve\_61-70, S\_111, Avg\_bonds, FCC, HCP, Avg\_bulk, Avg\_surf, R\_std, Std\_bonds, Curve\_21-30, Curve\_31-40, q6q6\_avg\_bulk, S\_110, S\_100, angle\_avg, q6q6\_avg\_surf, S\_311, q6q6\_avg\_total
    \item \textit{Experimental}: R\_max, FCC, DECA, T,  tau, time, S\_100, S\_111
\end{itemize}

\begin{figure}[h!]
     \centering
     \begin{subfigure}[t]{0.4\textwidth}
         \centering
         \includegraphics[width=\textwidth]{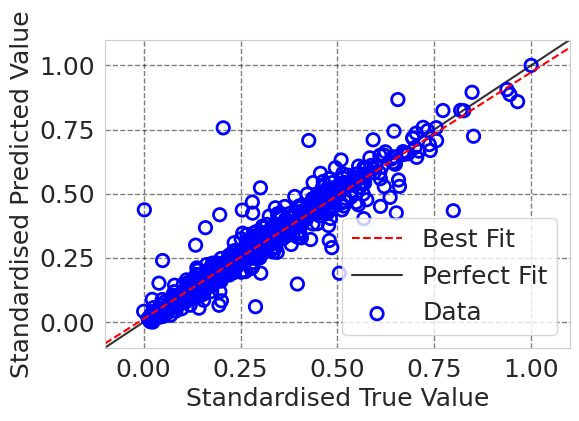}
         \caption{$R^2: 0.871 \pm 0.020$}
     \end{subfigure}
     \hfill
     \begin{subfigure}[t]{0.4\textwidth}
         \centering
         \includegraphics[width=\textwidth]{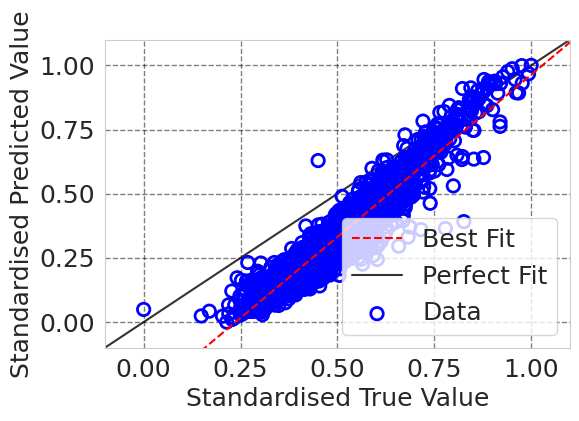}
          \caption{$R^2: 0.704 \pm 0.047$}
     \end{subfigure}
    \\ 
    \begin{subfigure}[t]{0.4\textwidth}
         \centering
         \includegraphics[width=\textwidth]{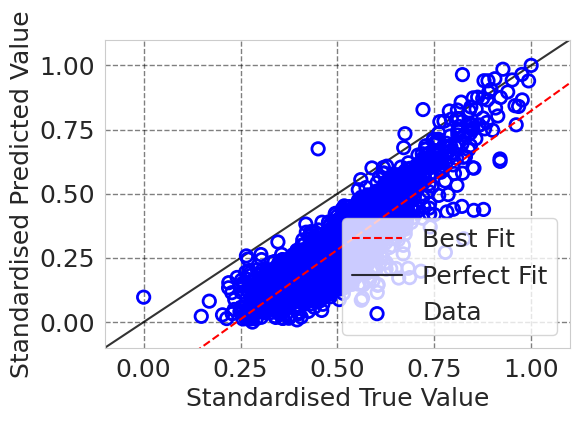}
          \caption{$R^2: 0.690 \pm 0.042$}
     \end{subfigure}
    \hfill
    \begin{subfigure}[t]{0.4\textwidth}
         \centering
         \includegraphics[width=\textwidth]{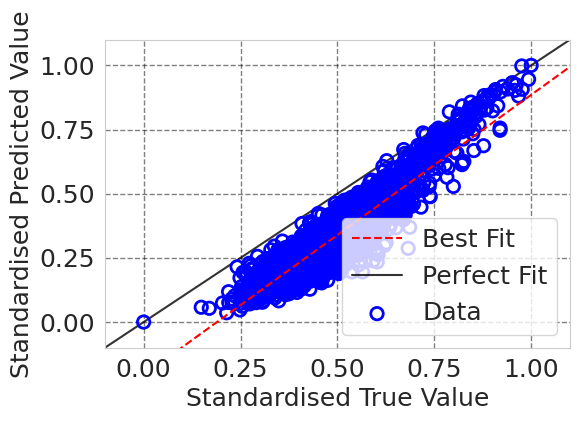}
          \caption{$R^2: 0.690 \pm 0.058$}
     \end{subfigure}
    \caption{
        The 45-degree plots comparing predicted versus actual values. (a) MLP for bulk modulus predictions, (b) RF for nanoparticle property prediction, (c) MLP for nanoparticle property prediction, and (d) XGBoost for nanoparticle property prediction}
    \label{fig:45-plots}
\end{figure}

We performed 5-fold cross-validation for all developed models and presented 45-degree plots with scores in Table \ref{fig:45-plots}.

\clearpage

\section*{Feature Descriptions For Datasets}
We provided the following Tables summarised feature descriptions used in the analysis.

\begin{scriptsize}
\begin{longtable}{lp{.75\textwidth}}
\toprule
\multicolumn{2}{l}{\textbf{Processing Features}}                                                                       \\ \midrule
T                & Temperature, K                                                                                      \\
tau              & Growth rate, atoms/ns                                                                               \\
time             & Time, ns                                                                                            \\ \midrule
\multicolumn{2}{l}{\textbf{Structural Features}}                                                                       \\ \midrule
N\_total         & Total number of atoms                                                                               \\
N\_bulk          & Total number of bulk atoms                                                                          \\
N\_surface       & Total number of surface atoms                                                                       \\
Volume           & Total nanoparticle volume, m3                                                                       \\
R\_min           & Nanoparticle radius minimum, Å                                                                      \\
R\_max           & Nanoparticle radius maximum,    Å                                                                   \\
R\_diff          & Nanoparticle radius minimum, Å                                                                      \\
R\_avg           & Nanoparticle radius average, Å                                                                      \\
R\_std           & Nanoparticle radius standard deviation, Å                                                           \\
R\_skew          & Nanoparticle radius skewness, Å                                                                     \\
R\_kurt          & Nanoparticle radius kurtosis, Å                                                                     \\
S\_100           & Number of atoms located on (100) surfaces                                                           \\
S\_111           & Number of atoms located on (111) surfaces                                                           \\
S\_110           & Number of atoms located on (110) surfaces                                                           \\
S\_311           & Number of atoms located on (311) surfaces                                                           \\
Curve\_1‐10      & Atoms with surface curvature angle between 1 and 10 degrees                                         \\
Curve\_11‐20     & Atoms with surface curvature angle between 11 and 20 degrees                                        \\
Curve\_21‐30     & Atoms with surface curvature angle between 21 and 30 degrees                                        \\
Curve\_31‐40     & Atoms with surface curvature angle between 31 and 40 degrees                                        \\
Curve\_41‐50     & Atoms with surface curvature angle between 41 and 50 degrees                                        \\
Curve\_51‐60     & Atoms with surface curvature angle between 51 and 60 degrees                                        \\
Curve\_61‐70     & Atoms with surface curvature angle between 61 and 70 degrees                                        \\
Curve\_71‐80     & Atoms with surface curvature angle between 71 and 80 degrees                                        \\
Curve\_81‐90     & Atoms with surface curvature angle between 81 and 90 degrees                                        \\
Curve\_91‐100    & Atoms with surface curvature angle between 91 and 100 degrees                                       \\
Curve\_101‐110   & Atoms with surface curvature angle between 101 and 110 degrees                                      \\
Curve\_111‐120   & Atoms with surface curvature angle between 111 and 120 degrees                                      \\
Curve\_121‐130   & Atoms with surface curvature angle between 121 and 130 degrees                                      \\
Curve\_131‐140   & Atoms with surface curvature angle between 131 and 140 degrees                                      \\
Curve\_141‐150   & Atoms with surface curvature angle between 141 and 150 degrees                                      \\
Curve\_151‐160   & Atoms with surface curvature angle between 151 and 160 degrees                                      \\
Curve\_161‐170   & Atoms with surface curvature angle between 161 and 170 degrees                                      \\
Curve\_171‐180   & Atoms with surface curvature angle between 171 and 180 degrees                                      \\
Avg\_total       & Order parameters, Average coordination number of all atoms                                          \\
Avg\_bulk        & Coordination statistics, Average coordination number of all   bulk atoms                            \\
Avg\_surf        & Coordination statistics, Average coordination number of all   surface atoms                         \\
TCN\_0           & Coordination statistics, Number of atoms with coordination   number 0                               \\
TCN\_1           & Coordination statistics, Number of atoms with coordination   number 1                               \\
TCN\_2           & Coordination statistics, Number of atoms with coordination   number 2                               \\
TCN\_3           & Coordination statistics, Number of atoms with coordination   number 3                               \\
TCN\_4           & Coordination statistics, Number of atoms with coordination   number 4                               \\
TCN\_5           & Coordination statistics, Number of atoms with coordination   number 5                               \\
TCN\_6           & Coordination statistics, Number of atoms with coordination   number 6                               \\
TCN\_7           & Coordination statistics, Number of atoms with coordination   number 7                               \\
TCN\_8           & Coordination statistics, Number of atoms with coordination   number 8                               \\
TCN\_9           & Coordination statistics, Number of atoms with coordination   number 9                               \\
TCN\_10          & Coordination statistics, Number of atoms with coordination   number 10                              \\
TCN\_11          & Coordination statistics, Number of atoms with coordination   number 11                              \\
TCN\_12          & Coordination statistics, Number of atoms with coordination   number 12                              \\
TCN\_13          & Coordination statistics, Number of atoms with coordination   number 13                              \\
TCN\_14          & Coordination statistics, Number of atoms with coordination   number 14                              \\
TCN\_15          & Coordination statistics, Number of atoms with coordination   number 15                              \\
TCN\_16          & Coordination statistics, Number of atoms with coordination   number 16                              \\
TCN\_17          & Coordination statistics, Number of atoms with coordination   number 17                              \\
TCN\_18          & Coordination statistics, Number of atoms with coordination   number 18                              \\
TCN\_19          & Coordination statistics, Number of atoms with coordination   number 19                              \\
TCN\_20          & Coordination statistics, Number of atoms with coordination   number 20                              \\
BCN\_0           & Coordination statistics, Number of bulk atoms with coordination   number 0                          \\
BCN\_1           & Coordination statistics, Number of bulk atoms with coordination   number 1                          \\
BCN\_2           & Coordination statistics, Number of bulk atoms with coordination   number 2                          \\
BCN\_3           & Coordination statistics, Number of bulk atoms with coordination   number 3                          \\
BCN\_4           & Coordination statistics, Number of bulk atoms with coordination   number 4                          \\
BCN\_5           & Coordination statistics, Number of bulk atoms with coordination   number 5                          \\
BCN\_6           & Coordination statistics, Number of bulk atoms with coordination   number 6                          \\
BCN\_7           & Coordination statistics, Number of bulk atoms with coordination   number 7                          \\
BCN\_8           & Coordination statistics, Number of bulk atoms with coordination   number 8                          \\
BCN\_9           & Coordination statistics, Number of bulk atoms with coordination   number 9                          \\
BCN\_10          & Coordination statistics, Number of bulk atoms with coordination   number 10                         \\
BCN\_11          & Coordination statistics, Number of bulk atoms with coordination   number 11                         \\
BCN\_12          & Coordination statistics, Number of bulk atoms with coordination   number 12                         \\
BCN\_13          & Coordination statistics, Number of bulk atoms with coordination   number 13                         \\
BCN\_14          & Coordination statistics, Number of bulk atoms with coordination   number 14                         \\
BCN\_15          & Coordination statistics, Number of bulk atoms with coordination   number 15                         \\
BCN\_16          & Coordination statistics, Number of bulk atoms with coordination   number 16                         \\
BCN\_17          & Coordination statistics, Number of bulk atoms with coordination   number 17                         \\
BCN\_18          & Coordination statistics, Number of bulk atoms with coordination   number 18                         \\
BCN\_19          & Coordination statistics, Number of bulk atoms with coordination   number 19                         \\
BCN\_20          & Coordination statistics, Number of bulk atoms with coordination   number 20                         \\
SCN\_0           & Coordination statistics, Number of surface atoms with   coordination number 0                       \\
SCN\_1           & Coordination statistics, Number of surface atoms with   coordination number 1                       \\
SCN\_2           & Coordination statistics, Number of surface atoms with   coordination number 2                       \\
SCN\_3           & Coordination statistics, Number of surface atoms with   coordination number 3                       \\
SCN\_4           & Coordination statistics, Number of surface atoms with   coordination number 4                       \\
SCN\_5           & Coordination statistics, Number of surface atoms with   coordination number 5                       \\
SCN\_6           & Coordination statistics, Number of surface atoms with   coordination number 6                       \\
SCN\_7           & Coordination statistics, Number of surface atoms with   coordination number 7                       \\
SCN\_8           & Coordination statistics, Number of surface atoms with   coordination number 8                       \\
SCN\_9           & Coordination statistics, Number of surface atoms with   coordination number 9                       \\
SCN\_10          & Coordination statistics, Number of surface atoms with   coordination number 10                      \\
SCN\_11          & Coordination statistics, Number of surface atoms with   coordination number 11                      \\
SCN\_12          & Coordination statistics, Number of surface atoms with   coordination number 12                      \\
SCN\_13          & Coordination statistics, Number of surface atoms with   coordination number 13                      \\
SCN\_14          & Coordination statistics, Number of surface atoms with   coordination number 14                      \\
SCN\_15          & Coordination statistics, Number of surface atoms with   coordination number 15                      \\
SCN\_16          & Coordination statistics, Number of surface atoms with   coordination number 16                      \\
SCN\_17          & Coordination statistics, Number of surface atoms with   coordination number 17                      \\
SCN\_18          & Coordination statistics, Number of surface atoms with   coordination number 18                      \\
SCN\_19          & Coordination statistics, Number of surface atoms with   coordination number 19                      \\
SCN\_20          & Coordination statistics, Number of surface atoms with   coordination number 20                      \\
Avg\_bonds       & Bonding statistics, Average bond length, Å                                                          \\
Std\_bonds       & Bonding statistics, Standard Deviation of the bond length, Å                                        \\
Max\_bonds       & Bonding statistics, Maximum bond length, Å                                                          \\
Min\_bonds       & Bonding statistics, Minimum bond length, Å                                                          \\
N\_bonds         & Bonding statistics, Total number of bonds                                                           \\
angle\_avg       & Bonding statistics, Average bond angle, Degrees                                                     \\
angle\_std       & Bonding statistics, Standard deviations of the bond angle,   Degrees                                \\
FCC              & Lattice statistics, Number of atoms in face centred cubic (fcc)   lattice                           \\
HCP              & Lattice statistics, Number of atoms in hexagonal closed packed   (hcp) lattice                      \\
ICOS             & Lattice statistics, Number of atoms in icosahedral lattice                                          \\
DECA             & Lattice statistics, Number of atoms in decahedral lattice                                           \\
q6q6\_avg\_total & Order parameters, Average spherical harmonic (q6.q6 \textgreater{}0.7)   for all atoms              \\
q6q6\_avg\_bulk  & Order parameters, Average spherical harmonic (q6.q6 \textgreater{}0.7)   for all bulk atoms         \\
q6q6\_avg\_surf &
  Order parameters, Average spherical harmonic (q6.q6 \textgreater{}0.7)   for all surface atoms \\
q6q6\_T0         & Order parameters, Number of atoms with spherical harmonic   (q6.q6 \textgreater{}0.7) of 0          \\
q6q6\_T1         & Order parameters, Number of atoms with spherical harmonic   (q6.q6 \textgreater{}0.7) of 1          \\
q6q6\_T2         & Order parameters, Number of atoms with spherical harmonic   (q6.q6 \textgreater{}0.7) of 2          \\
q6q6\_T3         & Order parameters, Number of atoms with spherical harmonic   (q6.q6 \textgreater{}0.7) of 3          \\
q6q6\_T4         & Order parameters, Number of atoms with spherical harmonic   (q6.q6 \textgreater{}0.7) of 4          \\
q6q6\_T5         & Order parameters, Number of atoms with spherical harmonic   (q6.q6 \textgreater{}0.7) of 5          \\
q6q6\_T6         & Order parameters, Number of atoms with spherical harmonic   (q6.q6 \textgreater{}0.7) of 6          \\
q6q6\_T7         & Order parameters, Number of atoms with spherical harmonic   (q6.q6 \textgreater{}0.7) of 7          \\
q6q6\_T8         & Order parameters, Number of atoms with spherical harmonic   (q6.q6 \textgreater{}0.7) of 8          \\
q6q6\_T9         & Order parameters, Number of atoms with spherical harmonic   (q6.q6 \textgreater{}0.7) of 9          \\
q6q6\_T10        & Order parameters, Number of atoms with spherical harmonic   (q6.q6 \textgreater{}0.7) of 10         \\
q6q6\_T11        & Order parameters, Number of atoms with spherical harmonic   (q6.q6 \textgreater{}0.7) of 11         \\
q6q6\_T12        & Order parameters, Number of atoms with spherical harmonic   (q6.q6 \textgreater{}0.7) of 12         \\
q6q6\_T13        & Order parameters, Number of atoms with spherical harmonic   (q6.q6 \textgreater{}0.7) of 13         \\
q6q6\_T14        & Order parameters, Number of atoms with spherical harmonic   (q6.q6 \textgreater{}0.7) of 14         \\
q6q6\_T15        & Order parameters, Number of atoms with spherical harmonic   (q6.q6 \textgreater{}0.7) of 15         \\
q6q6\_T16        & Order parameters, Number of atoms with spherical harmonic   (q6.q6 \textgreater{}0.7) of 16         \\
q6q6\_T17        & Order parameters, Number of atoms with spherical harmonic   (q6.q6 \textgreater{}0.7) of 17         \\
q6q6\_T18        & Order parameters, Number of atoms with spherical harmonic   (q6.q6 \textgreater{}0.7) of 18         \\
q6q6\_T19        & Order parameters, Number of atoms with spherical harmonic   (q6.q6 \textgreater{}0.7) of 19         \\
q6q6\_T20        & Order parameters, Number of atoms with spherical harmonic   (q6.q6 \textgreater{}0.7) of 20         \\
q6q6\_T20+ &
  Order parameters, Number of atoms with spherical harmonic   (q6.q6 \textgreater{}0.7) greater than 20 \\
q6q6\_B0         & Order parameters, Number of bulk atoms with spherical harmonic   (q6.q6 \textgreater{}0.7) of 0     \\
q6q6\_B1         & Order parameters, Number of bulk atoms with spherical harmonic   (q6.q6 \textgreater{}0.7) of 1     \\
q6q6\_B2         & Order parameters, Number of bulk atoms with spherical harmonic   (q6.q6 \textgreater{}0.7) of 2     \\
q6q6\_B3         & Order parameters, Number of bulk atoms with spherical harmonic   (q6.q6 \textgreater{}0.7) of 3     \\
q6q6\_B4         & Order parameters, Number of bulk atoms with spherical harmonic   (q6.q6 \textgreater{}0.7) of 4     \\
q6q6\_B5         & Order parameters, Number of bulk atoms with spherical harmonic   (q6.q6 \textgreater{}0.7) of 5     \\
q6q6\_B6         & Order parameters, Number of bulk atoms with spherical harmonic   (q6.q6 \textgreater{}0.7) of 6     \\
q6q6\_B7         & Order parameters, Number of bulk atoms with spherical harmonic   (q6.q6 \textgreater{}0.7) of 7     \\
q6q6\_B8         & Order parameters, Number of bulk atoms with spherical harmonic   (q6.q6 \textgreater{}0.7) of 8     \\
q6q6\_B9         & Order parameters, Number of bulk atoms with spherical harmonic   (q6.q6 \textgreater{}0.7) of 9     \\
q6q6\_B10        & Order parameters, Number of bulk atoms with spherical harmonic   (q6.q6 \textgreater{}0.7) of 10    \\
q6q6\_B11        & Order parameters, Number of bulk atoms with spherical harmonic   (q6.q6 \textgreater{}0.7) of 11    \\
q6q6\_B12        & Order parameters, Number of bulk atoms with spherical harmonic   (q6.q6 \textgreater{}0.7) of 12    \\
q6q6\_B13        & Order parameters, Number of bulk atoms with spherical harmonic   (q6.q6 \textgreater{}0.7) of 13    \\
q6q6\_B14        & Order parameters, Number of bulk atoms with spherical harmonic   (q6.q6 \textgreater{}0.7) of 14    \\
q6q6\_B15        & Order parameters, Number of bulk atoms with spherical harmonic   (q6.q6 \textgreater{}0.7) of 15    \\
q6q6\_B16        & Order parameters, Number of bulk atoms with spherical harmonic   (q6.q6 \textgreater{}0.7) of 16    \\
q6q6\_B17        & Order parameters, Number of bulk atoms with spherical harmonic   (q6.q6 \textgreater{}0.7) of 17    \\
q6q6\_B18        & Order parameters, Number of bulk atoms with spherical harmonic   (q6.q6 \textgreater{}0.7) of 18    \\
q6q6\_B19        & Order parameters, Number of bulk atoms with spherical harmonic   (q6.q6 \textgreater{}0.7) of 19    \\
q6q6\_B20        & Order parameters, Number of bulk atoms with spherical harmonic   (q6.q6 \textgreater{}0.7) of 20    \\
q6q6\_B20+ &
  Order parameters, Number of bulk atoms with spherical harmonic   (q6.q6 \textgreater{}0.7) greater than 20 \\
q6q6\_S0         & Order parameters, Number of surface atoms with spherical   harmonic (q6.q6 \textgreater{}0.7) of 0  \\
q6q6\_S1         & Order parameters, Number of surface atoms with spherical   harmonic (q6.q6 \textgreater{}0.7) of 1  \\
q6q6\_S2         & Order parameters, Number of surface atoms with spherical   harmonic (q6.q6 \textgreater{}0.7) of 2  \\
q6q6\_S3         & Order parameters, Number of surface atoms with spherical   harmonic (q6.q6 \textgreater{}0.7) of 3  \\
q6q6\_S4         & Order parameters, Number of surface atoms with spherical   harmonic (q6.q6 \textgreater{}0.7) of 4  \\
q6q6\_S5         & Order parameters, Number of surface atoms with spherical   harmonic (q6.q6 \textgreater{}0.7) of 5  \\
q6q6\_S6         & Order parameters, Number of surface atoms with spherical   harmonic (q6.q6 \textgreater{}0.7) of 6  \\
q6q6\_S7         & Order parameters, Number of surface atoms with spherical   harmonic (q6.q6 \textgreater{}0.7) of 7  \\
q6q6\_S8         & Order parameters, Number of surface atoms with spherical   harmonic (q6.q6 \textgreater{}0.7) of 8  \\
q6q6\_S9         & Order parameters, Number of surface atoms with spherical   harmonic (q6.q6 \textgreater{}0.7) of 9  \\
q6q6\_S10        & Order parameters, Number of surface atoms with spherical   harmonic (q6.q6 \textgreater{}0.7) of 10 \\
q6q6\_S11        & Order parameters, Number of surface atoms with spherical   harmonic (q6.q6 \textgreater{}0.7) of 11 \\
q6q6\_S12        & Order parameters, Number of surface atoms with spherical   harmonic (q6.q6 \textgreater{}0.7) of 12 \\
q6q6\_S13        & Order parameters, Number of surface atoms with spherical   harmonic (q6.q6 \textgreater{}0.7) of 13 \\
q6q6\_S14        & Order parameters, Number of surface atoms with spherical   harmonic (q6.q6 \textgreater{}0.7) of 14 \\
q6q6\_S15        & Order parameters, Number of surface atoms with spherical   harmonic (q6.q6 \textgreater{}0.7) of 15 \\
q6q6\_S16        & Order parameters, Number of surface atoms with spherical   harmonic (q6.q6 \textgreater{}0.7) of 16 \\
q6q6\_S17        & Order parameters, Number of surface atoms with spherical   harmonic (q6.q6 \textgreater{}0.7) of 17 \\
q6q6\_S18        & Order parameters, Number of surface atoms with spherical   harmonic (q6.q6 \textgreater{}0.7) of 18 \\
q6q6\_S19        & Order parameters, Number of surface atoms with spherical   harmonic (q6.q6 \textgreater{}0.7) of 19 \\
q6q6\_S20        & Order parameters, Number of surface atoms with spherical   harmonic (q6.q6 \textgreater{}0.7) of 20 \\
q6q6\_S20+ &
  \begin{tabular}[c]{@{}l@{}}Order parameters, Number of surface atoms with spherical   harmonic (q6.q6 \textgreater{}0.7) greater\\      than 20\end{tabular} \\ \midrule
\multicolumn{2}{l}{\textbf{Target Property Labels}}                                                                    \\ \midrule
Total\_E         & Total energy of the nanoparticle from the LAMMPS simulation, eV                                     \\
Formation\_E &
  Formation energy of the nanoparticle (Total\_E‐   N\_total*Bulk\_E/atom), eV Where the Bulk\_E/atom is provided on the website for   the EAM potential \\ \bottomrule

\caption{Metallic Nanoparticle Header List}
\end{longtable}
\end{scriptsize}
\begin{table}[ph!]
\centering
\caption{Features and Labels Description of BACE-1 Classification Case Study}
\label{tab:my-table}
\begin{tabular}{lll}
\toprule
       & Type   & Description                   \\ \midrule
SMILES & String & Physicochemical Compounds     \\
Label  & Int    & Indicating Active or Inactive \\ \bottomrule
\end{tabular}
\end{table}
\begin{table}[ph!]
\centering
\caption{Features and Labels Description of AFLOW Prediction Case Study}
\label{tab:my-table}
\begin{tabular}{lll}
\toprule
       & Type   & Description                   \\ \midrule
Formula & String & Compounds     \\
Target  & Float  & Bulk modulus value \\ \bottomrule
\end{tabular}
\end{table}

\setcounter{Maxaffil}{0}
\renewcommand\Affilfont{\itshape\small}

\clearpage

\begin{figure}[!h]
\centerline{\includegraphics[width=\columnwidth]{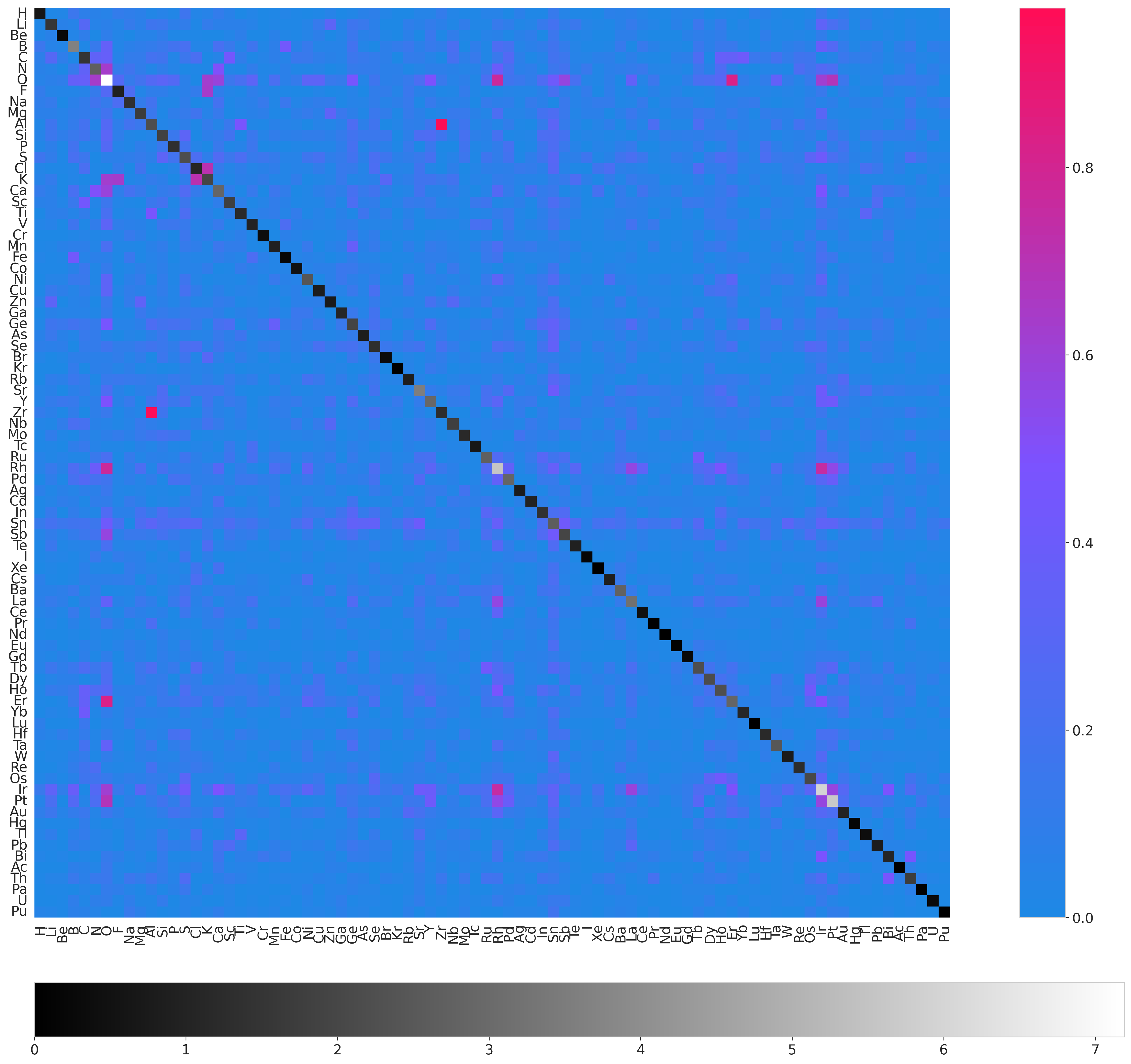}}
\caption{Second-order feature interaction attribution calculated based on the well-trained MLP. The diagonal is coloured in a grayscale gradient, reflecting increasing first-order attributions. The red-blue heatmap illustrates the absolute interaction strength: red-coloured cells indicate stronger attributions, while blue-coloured ones indicate weaker interactions.}
\label{fig:aflow-interaction}
\end{figure}

\clearpage
\begin{figure}[p!]
     \centering
     \includegraphics[width=\textwidth]{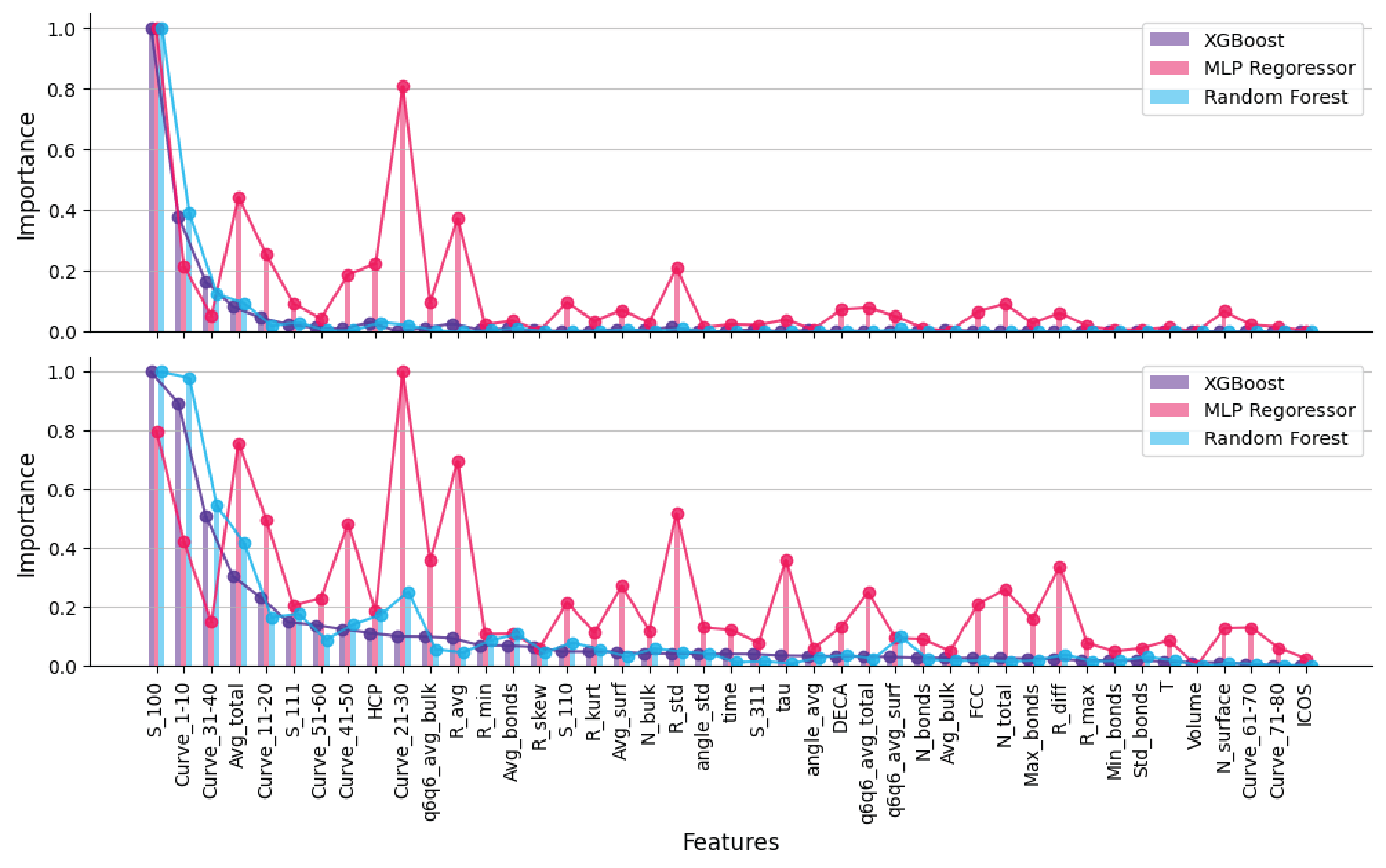}
\caption{\label{fig:nano-model-outcomes}Feature importance rankings of metallic nanoparticles from accurate models, including XGBoost, MLP, and RF and well-established explanation methods, including Shap, PI, and IG. Feature importance rankings (SHAP) from different well-trained models (top), and feature importance rankings (PI) from different well-trained models (bottom). The x-axis displays features ordered by their importance ranking from RF, which serves as a baseline. Other rankings are plotted according to this order. The y-axis denotes the importance score, normalised between 0 and 1.}
\end{figure}

\begin{figure}[h!]
     \centering
         \includegraphics[width=\textwidth]{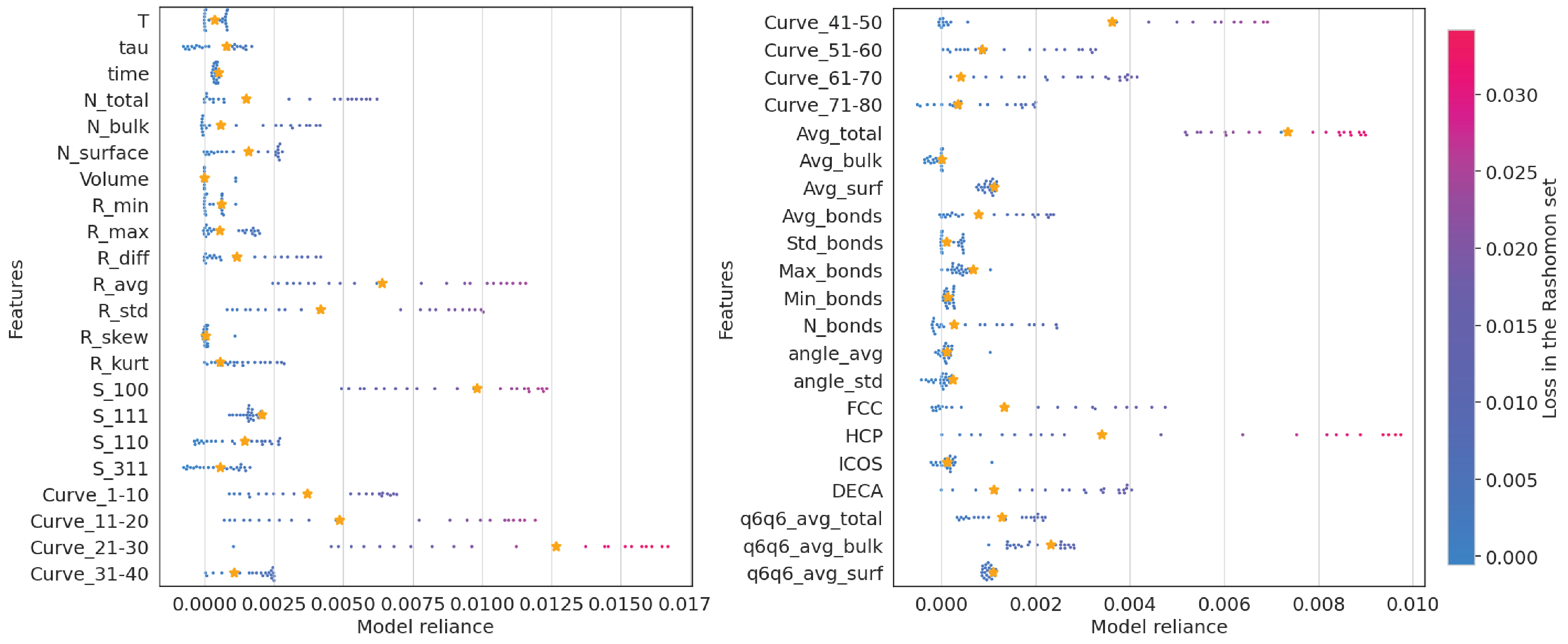}
         \label{fig:fig3}
    \caption{
    Illustration of the first-order explanation for a set of well-trained models for the task of nanoparticle fractal dimension prediction (noting well-trained does not imply high predictive performance). The yellow star represents the reference feature importance, and each point is colored according to its loss. Model reliance is the term in the literature \cite{fisher2019all}, meaning feature importance in this study} 
\label{fig:nano-rset}
\end{figure}
